\documentclass[a4paper,fleqn]{cas-dc}

\usepackage{xcolor}
\hypersetup{
 colorlinks,
linkcolor={black},
}

\usepackage{enumitem}
\setlist{nosep}

\usepackage[authoryear]{natbib}
\usepackage[acronym]{glossaries}
\usepackage{algorithm}
\usepackage{amssymb}
\usepackage{color,soul}
\usepackage{float}
\usepackage{graphicx}
\usepackage{listings}
\usepackage{subfigure}
\usepackage{tabularx}
\usepackage{array}
\usepackage{multirow}
\usepackage{multicol}
\usepackage{cleveref}

\newcolumntype{L}[1]{>{\raggedright\let\newline\\\arraybackslash\hspace{0pt}}m{#1}}
\newcolumntype{R}[1]{>{\hsize=#1\hsize\raggedright\arraybackslash}X}%

\begin{document}
\newacronym{uav}{UAV}{Unmanned Aerial Vehicle}
\newacronym{ugv}{UGV}{Unmanned Ground Vehicle}
\newacronym{slam}{SLAM}{Simultaneous Localisation and Mapping}
\newacronym{ga}{GA}{Genetic Algorithm}
\newacronym{ge}{GE}{Grammatical Evolution}
\newacronym{ep}{EP}{Evolutionary Programming}
\newacronym{ann}{ANN}{Artifical Neural Network}
\newacronym{hh}{HH}{Hyper-Heuristics}
\newacronym{rhs}{RHS}{Right-Hand Side}
\newacronym{lhs}{LHS}{Left-Hand Side}
\newacronym{ABC}{ABC}{Artificial Bee Colony}
\newacronym{pso}{PSO}{Particle Swarm Optimisation}
\newacronym{wsn}{WSN}{Wireless Sensor Network}
\newacronym{ai}{AI}{Artificial Intelligence}
\newacronym{ROS}{ROS}{Robot Operating System}
\newacronym{RTS/CTS}{RTS/CTS}{Request-to-send/ Clear-to-send}
\newacronym{fsm}{FSM}{Finite State Machine}
\newacronym{manet}{MANET}{Mobile Ad-hoc NETwork}
\newacronym{gn}{GN}{Graph Neuron}
\newacronym{dhgn}{DHGN}{Distributed Hierarchical Graph Neurons}
\newacronym{ml}{ML}{Machine Learning}
\newacronym{lcs}{LCS}{Learning Classifier System}
\newacronym{bba}{BBA}{Bucket Brigade Algorithm}
\newacronym{TCS}{TCS}{Temporal Classifier System}
\newacronym{ZCS}{ZCS}{Zeroth-level Classifier System}
\newacronym{llh}{LLH}{Low-Level Heuristics}
\newacronym{XCS}{XCS}{eXended Classifier System}
\newacronym{SAMUEL}{SAMUEL}{Strategy Acquisition Method Using Empirical Learning}
\newacronym{SARSA}{SARSA}{State-Action, Reward, State-Action}
\newacronym{rl}{RL}{Reinforcement Learning}
\newacronym{GRASP}{GRASP}{Greedy Randomised Adaptive Search Procedures}
\newacronym{WOLF}{WOLF}{Win Or Learn Fast}
\newacronym{DSTG}{DSTG}{Defence Science and Technoogy Group}
\newacronym{ldpl}{LDPL}{Log-Distance Path Loss}
\newacronym{snr}{SNR}{Signal to Noise Ratio}
\newacronym{snir}{SNIR}{Signal to Noise and Interference Ratio}
\newacronym{csma}{CSMA}{Carrier-Sense Multiple Access}
\newacronym{uct}{UCT}{Upper Confidence Bound applied to trees}
\newacronym{rnn}{RNN}{Recurrent Neural Network}
\newacronym{dnn}{DNN}{Deep Neural Network}
\newacronym{hgn}{HGN}{Hierarchical Graph Neurons}
\newacronym{rhgn}{R-HGN}{Robotic Hierarchical Graph Neurons}

\newacronym{hhgn}{$\mathrm{H}^2$GN}{Hierarchical, Hierarchical Graph Neurons}
\newacronym{mlp}{MLP}{Multi-Layer Perceptron}

\newacronym{CNN}{CNN}{Convolutional Neural Network}
\newacronym{enn}{ENN}{Evolved Neural Network}
\newacronym{snn}{SNN}{Supervised training Neural Network}
\newacronym{ci}{CI}{Confidence Interval}
\newacronym{mb}{MB}{Manually designed Behaviour}
\newacronym{tof}{ToF}{Time-of-Flight}
\newacronym{tdma}{TDMA}{Time-division Multiple Access}
\newacronym{fcc}{FCC}{Federal Communications Commission}
\newacronym{ta}{TA}{Threshold Acceptance}
\newacronym{ce}{CE}{Conformit\'{e} Europ\'{e}ene}
\newacronym{SINR}{SINR}{Signal to Interference, plus Noise Ratio}
\newacronym{des}{DES}{Discrete Event Simulation}
\newacronym{Nov}{NDS}{Novelty Domination Search}
\newacronym{long}{long}{long range}
\newacronym{short}{short}{short range}

\let\WriteBookmarks\relax
\def\floatpagepagefraction{1}
\def\textpagefraction{.001}
\shorttitle{RHGN. A novel implementation in swarm robotic behaviour control}
\shortauthors{P. Smith, \textit{et~al.}}

\title [mode = title]{Robotic Hierarchical Graph Neurons. A novel implementation of HGN for swarm robotic behaviour control}

\author[1]{Phillip~Smith}[orcid=0000-0002-5330-1830]
\cormark[1]
\credit{Software, Conducted Simulations, Interpreted Results, Writing - Original draft preparation}

\author[1]{Aldeida~Aleti}\credit{Project Supervisor, Writer - feedback}
\author[1]{Vincent~C.S.~Lee}\credit{Project Supervisor, Writer - feedback}
\author[2]{Robert~Hunjet}\credit{Project Supervisor, Writer - feedback,Conceptualization of this study}
\author[3]{Asad~Khan}\credit{Project Supervisor, Writer - feedback,Conceptualization of this study}

\address[1]{Faculty of Information Technology, Monash University, 25 Exhibition Walk, Clayton, Victoria, Australia, 3168}
\address[2]{Defence Science and Technology Group, PO Box 1500, Edinburgh, South Australia, Australia, 5111}
\address[3]{Sensor Analytics, PO Box 388, Mount Waverley, Victoria, Australia, 3149}

\cortext[cor1]{Corresponding author}

\begin{abstract}
This paper explores the use of a novel form of Hierarchical Graph Neurons (HGN) for in-operation behaviour selection in a swarm of robotic agents. This new HGN is called Robotic-HGN (R-HGN), as it matches robot environment observations to environment labels via fusion of match probabilities from both temporal and intra-swarm collections. This approach is novel for HGN as it addresses robotic observations being pseudo-continuous numbers, rather than categorical values. Additionally, the proposed approach is memory and computation-power conservative and thus is acceptable for use in mobile devices such as single-board computers, which are often used in mobile robotic agents. This R-HGN approach is validated against individual behaviour implementation and random behaviour selection. This contrast is made in two sets of simulated environments: environments designed to challenge the held behaviours of the R-HGN, and randomly generated environments which are more challenging for the robotic swarm than R-HGN training conditions. R-HGN has been found to enable appropriate behaviour selection in both these sets, allowing significant swarm performance in pre-trained and unexpected environment conditions. 
\end{abstract}


\maketitle
\let\thefootnote\relax\footnotetext{Email addresses: phillipsmith@monash.edu (P.~Smith),  aldeida.aleti@monash.edu (A.~Aleti), Vincent.CS.Lee@monash.edu (CS.~Lee), Robert.Hunjet@dsto.defence.gov.au (R.~Hunjet), asad.khan@sensoranalytics.com.au (A.~Khan)}

\let\thefootnote\relax\footnotetext{Abbreviations: Env.= Environment }
\section{Introduction}\label{intro}
Recent advancements in swarm robotic behaviour creation has seen an increase in task effectiveness for swarm robotics in non-trivial tasks, such as data-transfer via \glspl{manet} creation \citep{smith2018swarm}.
However, this behaviour evolution approach suffers from poor transferability between environments and problem instances. That is, significant variation between evolution environment(s) and operation environment(s) sees considerable reductions in operation task performance.

To overcome this limitation, this paper explores swarm behaviour macro-adjustments in the form of environment identification for behaviour switching. We achieve this switch by training a \gls{hgn} pattern classifier with environment observations and associating each pattern with a behaviour from a pre-defined repertoire.  The use of \gls{hgn} has been selected due to its noted ability to work on computationally-limited devices, such as single board computers used for swarm robotics \citep{nasution2008hierarchical}. However, as environmental data often contains mixed inputs which may including pseudo-continuous numbers, rather than uniform categorical values, and as local environment observations may relate to multiple global environments, this paper presents a novel extension to \gls{hgn}, \gls{rhgn}.

The exploration of this \gls{rhgn} in swarm behaviour switching is conducted in this paper via simulations of a data-transfer task seen in \cite{smith2018swarm}. This task is partially-observable, and requires behaviour heterogeneity both across the swarm and across task duration. As such, it is seen as a significant test-bed for the \gls{rhgn} swarm implementation. 

Two experiments are presented in this paper to explore the \gls{rhgn} swarm: 
pre-trained environment operations and untrained environment operations. 
The former validates the \gls{rhgn} implementation by training the \gls{rhgn} in a collection of environments specifically developed for the behaviour repertoire, and evaluating in these same environments or in concatenations of these environments. The latter implements the \gls{rhgn} driven swarm in randomly generated environments without further training. Thus emulating the swarm being deployed in an unpredictable operation with no specific prior preparation. The \gls{rhgn} is evaluated in terms of resulting swarm performance compared to each behaviour being solitarily utilised, and against a random behaviour selector. Additionally, the environment matching accuracy of \gls{rhgn} is assessed.

The primary contribution of this paper is \textbf{a novel implementation of \gls{hgn} for environment matching in robotic agents, \gls{rhgn}}.
This contribution is achieved via altering the pattern memory structure of the \glspl{gn}  and by altering the output to probabilistic environment-matches which utilise temporal prediction fusion for associated behaviour selection. Additionally, as this \gls{rhgn} is implemented in a swarm of robots, prediction fusion is also conducted via intra-swarm sharing.
The value of this novel \gls{hgn} implementation is determined via:
exploration of the performance by  an \gls{rhgn} equipped robotic swarm in both environments known \textit{a priori} and unknown; analysis of \gls{rhgn} pattern matching accuracy; and comparison of the above qualities against a swarm randomly selecting behaviours.

\section{Related work}\label{lit}

\subsection{Behaviour Selection}
In this study, robot behaviour selection takes heavy inspiration from the growing research field of \gls{hh} \citep{Burke2013, Kochenberger2003}. This inspiration draws directly from \gls{hh} as `heuristics' and `behaviours' are seen to be interchangeable ideas across different disciplines. 

In \gls{hh}, a wide array of meta-heuristics, data-mining and meta-learning algorithms have been implemented to have a system utilise an appropriate heuristic for each problem instance. This heuristic selection has been either a single decision during start-up \citep{ tabataba2012,nagavalli2017automated,leng2017task, burke2006case, Thabtah2008, terashima2008, smith2008towards, hagenauer2017comparative} or a periodic decision during operation \citep{misir2009,tavares2018algorithms,soria2014}.

For small-scale heuristic selection, a trial and error approach was presented in \cite{tabataba2012}. Each heuristic was tested in a limited run simulation and the best (known) heuristic utilised in a complete implementation. This approach was limited to a selection during startup and had computation requirements scaling $O(n)$ from heuristics stored. Additionally, behaviours with deceptive early-operation performance may have impacted the selection quality.
This simulation trialling was also seen in the field of swarm robotics in \cite{nagavalli2017automated} with heuristic selection being used for A* swarm behaviour sequence searching. This approach allowed for complete behaviour planning in homogeneous swarm tasks. However, it was limited to deterministic tasks, such as targeted locomotion or area coverage. Additionally, this work relied on a centralised controller being supplied all swarm agent observations and commanded all swarm agent behaviour switches.

To avoid the drawbacks of the trial-and-error approaches, permanent knowledge may be created which associates novel problem instances and pseudo optimal heuristics \citep{ burke2006case, Thabtah2008, terashima2008, smith2008towards,hagenauer2017comparative}. 
In \cite{leng2017task} a centralised swarm distribution system utilised such knowledge. This system held associations between required tasks and swarm behaviours. However, the agents of this system did not autonomously derive the required task from environment observations but rather relied on operator commands.  In contrast, \cite{burke2006case} presented a selection method which derived an optimal feature vector for problem-heuristic associations. Similarly, in \cite{smith2008towards} a feed-forward \gls{mlp} neural network was trained for optimal meta-heuristic prediction via problem feature inputs. Finally, in \cite{Thabtah2008} and \cite{hagenauer2017comparative} problems were categorised via decision trees created from past experiences, with \cite{hagenauer2017comparative} finding these trees to be more accurate than both feature vectors and \gls{mlp}. However, all these approaches were limited to  single behaviour/heuristic selection process during start-up. Such an approach is known to limit overall accuracy when operating in partially observable environments, as initial environmental observations may not be indicative of the environment as a whole \cite{smallwood1973optimal}. 

The variance in algorithms used in \gls{hh} selection techniques, along with the similarities drawn between \textit{heuristic selection} and all meta-algorithms in \cite{Pappa2014}, show that heuristic selection may be achieved by any pattern matching techniques. This is particularly true for large, pseudo-continuous value, pattern matching, which, for many of the above algorithms, is known to reduce recall accuracy \citep{kim2008comparison} and require larger training sets \citep{baum1989size} and time \citep{hettiarachchige2018multi}. Additionally, these algorithms are seen to suffer from poor recall accuracy as pattern similarities increase \citep{nasution2008hierarchical}.  To address these shortcomings, this research looks to an alternative pattern matching algorithm, \gls{hgn}.

\subsection{Hierarchical Graph Neurons}
In 2008, the one-shot learning algorithm, \gls{hgn}, was developed by \cite{nasution2008hierarchical}. This algorithm expanded on the original, single layer, graph neuron algorithm \citep{khan2002peer}, overcoming the \textit{crosstalk} issue. In this debut paper \gls{hgn} was seen to achieve pseudo-real-time recall speeds; accurately recalling patterns with discrete identifications, even with significant pattern distortion;  and was noted to be applicable for small, single-board computers, such as wireless sensor modules. Additionally, \gls{hgn} was compared to a single-cycle back-propagation \gls{mlp} which demonstrated a significant superiority by \gls{hgn} in regards to scaling impact, both for pattern size and memorised quantity.

As a brief overview , \gls{hgn} is a layered architecture of \glspl{gn} for pattern classification. The base layer consists of a \gls{gn} row for each pattern component and a \gls{gn} column for each value the components may take. Each subsequent layer of the \gls{hgn} has two fewer rows than the previous, with the outer-most rows being removed. This layered reduction continues until a one-row layer is created as the top of the triangular hierarchy. When a pattern is passed to the \gls{hgn}  one \gls{gn} of each row is activated in the base layer, with the activation being determined by the component value and the associated column index. The activated \glspl{gn} broadcast the pattern value to all \glspl{gn} in neighbouring rows and in-turn receives two value broadcasts. The combination of the received values is then associated with a sub-pattern identification in each active \gls{gn} and passed up to the \gls{gn} with equal row and column position in the above layer. This sub-pattern recalling is repeated in each layer until a \gls{gn} in the top-layer is activated. This top-layer recall is the combination of all prior sub-pattern recalls and is thus representative of the entire pattern.

In the last decade \gls{hgn} has been specialised and refined for a range of problems. 
In  \cite{dhgn}, \gls{hgn} was refined to \gls{dhgn}. This improvement separated the input pattern into sub-patterns, each being individually classified via a \gls{hgn}. The overall pattern was classified via a majority vote. 
\gls{dhgn} permitted equivalent performance to \gls{hgn} with fewer computation neurons. As such, it was tested with distorted image identification, showing greater classification speed in comparison to a state-of-the-art algorithm of the time.  Finally, within the domain of swarm robotics, \cite{hettiarachchige2018multi} used a sequential \gls{hgn} for anti-swarm motion tracking. This work saw \gls{hgn} outperform a recurrent neural network.  

From this review, it can be concluded that \gls{hgn}, and more specifically \gls{dhgn}, is a light-weight, accurate and scalable pattern matching approach. \gls{hgn} shows potential for use in low-complexity agents, such as that used for swarm robots.

\section{System design}

In this section, the structure and novelties of the proposed \gls{rhgn} are presented and the training process for the explored swarm application is discussed. The \gls{rhgn} of this study utilises the distributed structure of \gls{dhgn} for the aforementioned computation reductions. However, the novel contributions of this algorithm are equally applicable to standard \gls{hgn}.

\subsection{Robotic \gls{hgn}}
The novel contributions of \gls{rhgn} target two identified issues with \gls{hgn} (or \gls{dhgn}). Firstly, the mixed inputs, which may contain pseudo-continuous numbers, taken from environment observations result in numerous unused pattern components. This results in standard (D)HGN  unnecessarily consuming considerable memory. Secondly, standard (D)\gls{hgn} output discrete pattern classifications, however, localised environment observations may be seen across multiple environments and thus requires a fuzzy pattern match.

\subsubsection{Mixed, pseudo-continuous inputs}
To further describe the first issue, let us explore a simple environment pattern with three inputs, \textit{A}, \textit{B} and \textit{C}, with respective ranges $\mathbf{Z}\in [0,10)$, $\mathbf{Z}$, and $\mathbf{Z}\in [0,1]$.
Traditional \gls{hgn} classify patterns consisting of uniform ranged categorical components and thus each column  will have a set \glspl{gn} of size \textit{r}. Additionally, the triangular structure of \gls{hgn} results in the total number of \gls{gn} being $ 
r \cdot\{n+(n-2)+(n-4)+\dots+1\} = r\cdot (n+1)^2/4$
 , where \textit{n} is the pattern length. 

\begin{figure}
	\centering
	\subfigure[Standard HGN, muliple redundant GN and connections.]{    \includegraphics[width=0.35\linewidth]{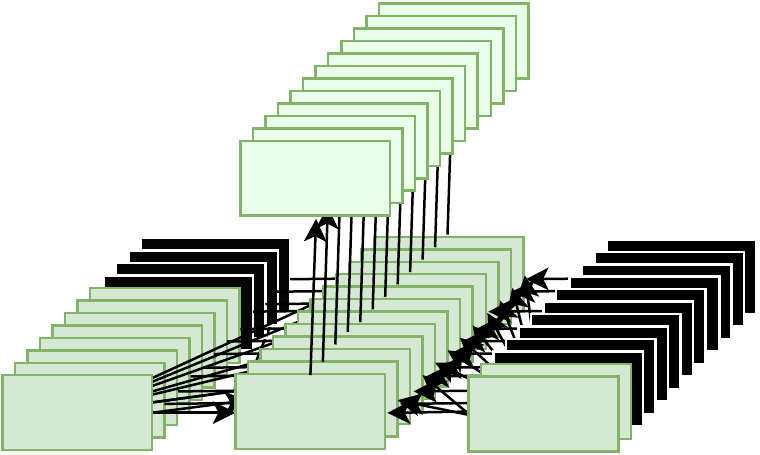}}%
		\hspace*{0.1\linewidth}
	\subfigure[\gls{rhgn}, only required GN created. One connecton per row.]{      \includegraphics[width=0.4\linewidth]{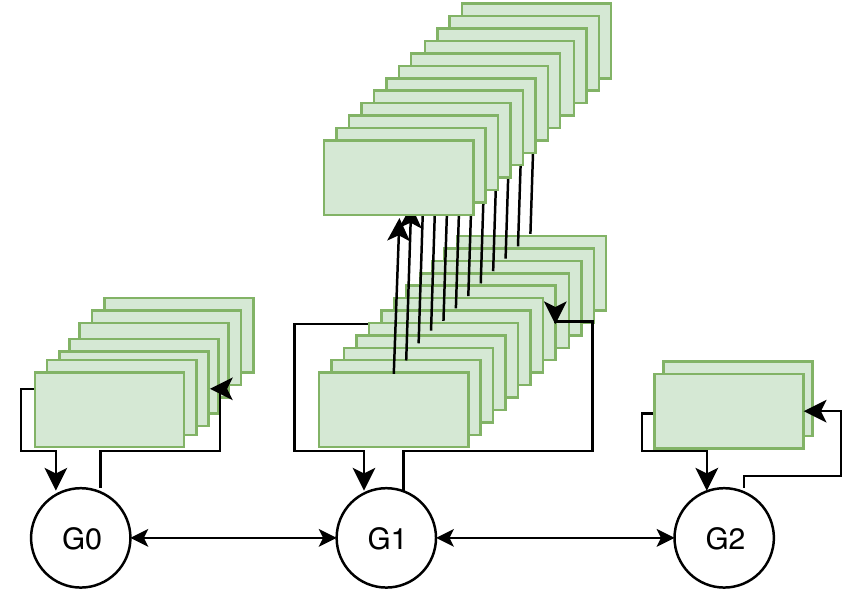}}
	\caption{Green/light boxes are utilised GN, black boxes are non-activatable. White circles are the row gates connected to neighbouring gates.  }
	\label{fig:hgn-digram-1}
\end{figure}

For mixed-range inputs, \textit{r} must be the largest value to accommodate the input ranges of all inputs, thus for this example $r=max(|A|,|B|,|C|) = |\mathbf{Z}|$ and therefore the \gls{hgn} is theoretically of  infinite size. However, as these observations will be made by robotic devices only pseudo-continuous ranges are possible. If we assume the agent is using unsigned 32-bit integers as $\mathbf{Z}$, \textit{r} becomes $4,294,967,295$ and a total of $17 \times 10^9$ \glspl{gn} are created during start-up, though only a fraction of these \glspl{gn} are utilised. That is, input \textit{C} has $\sim 4\times10^9$ \glspl{gn}, with only two being activatable.  A diagram of this traditional \gls{hgn} is shown in Figure~\ref{fig:hgn-digram-1} \textit{(a)}, with black squares being non-activatable \gls{gn}. 

A novice solution to this scaling issue is to create each \gls{hgn} row via individual variable range. However, such a solution will only reduce this example to $8.6 \cdot 10^9$  \glspl{gn}. Instead, \gls{rhgn} overcomes this issue by dynamically generating \glspl{gn} as each variable is observed in training. During this generation process a new \gls{gn} is added at the column of each layer at the allocated row. Additionally, to prevent invalid neighbour communication after row additions, each \glspl{gn} row is isolated behind a `gate'. These gates collect outbound messages from the active \glspl{gn} to pass to the neighbouring row's gate. The gates then send inbound messages to only these active \gls{gn}, rather than broadcasting to all \glspl{gn} in the row. An example of this dynamic GN creation and gated row connection is  shown in Figure~\ref{fig:hgn-digram-1}\textit{(b)}. Returning to the  sample environment pattern of \textit{A, B} and \textit{C}, we see a linear relation between observed component values and \glspl{gn} size. If we assume training consists of 2 million patterns, and all patterns have a unique value for \textit{B} (an unrealistic extreme) \gls{rhgn} will hold only  $4\times 10^6$ \glspl{gn}.
\subsubsection{Probability environment matches}\label{dynamciGN}
The second issue with (D)HGN is the agent localised observation patterns may correlate to many environments, while \gls{hgn} traditionally outputs a single pattern identification.  To overcome this issue, \gls{rhgn} associates each  discrete pattern classification to a probability tuple, where each probability relates to a possible environment.  To form these  probability outcomes, the training process of \gls{hgn} is extended to record the occurrence of each pattern identification for each environment. After all training data has been examined, these environment counts are normalised and stored in hash-tables using the unique pattern classification as the recall key. This process does not interfere with the underlying classification of \gls{hgn}, and thus a one-to-one relation is assured between each pattern and probability tuple.

In addition to these probabilities, \gls{rhgn} adjusts the upper sub-pattern concatenation of \gls{dhgn}. In standard \gls{dhgn}, after each sub-\gls{hgn} of the \gls{dhgn} classified a sub-pattern, a majority vote was used to determine the overall pattern. This approach introduced inaccuracy as significant variation in a  sub-pattern was overlooked during voting. In contrast, \gls{rhgn} implements an upper \gls{hgn}, taking the \textit{argmax} of the lower \glspl{hgn} probabilities as inputs. This allows variation in any one sub-pattern to be acknowledged during final pattern classification. Should this upper \gls{hgn} fail to match an observed pattern, an averaging decision fusion of the lower \gls{hgn} probability tuples is taken. 

\subsection{Swarm Environment Matching}     
\begin{figure}
	\centering
	\includegraphics[width=.8\linewidth]{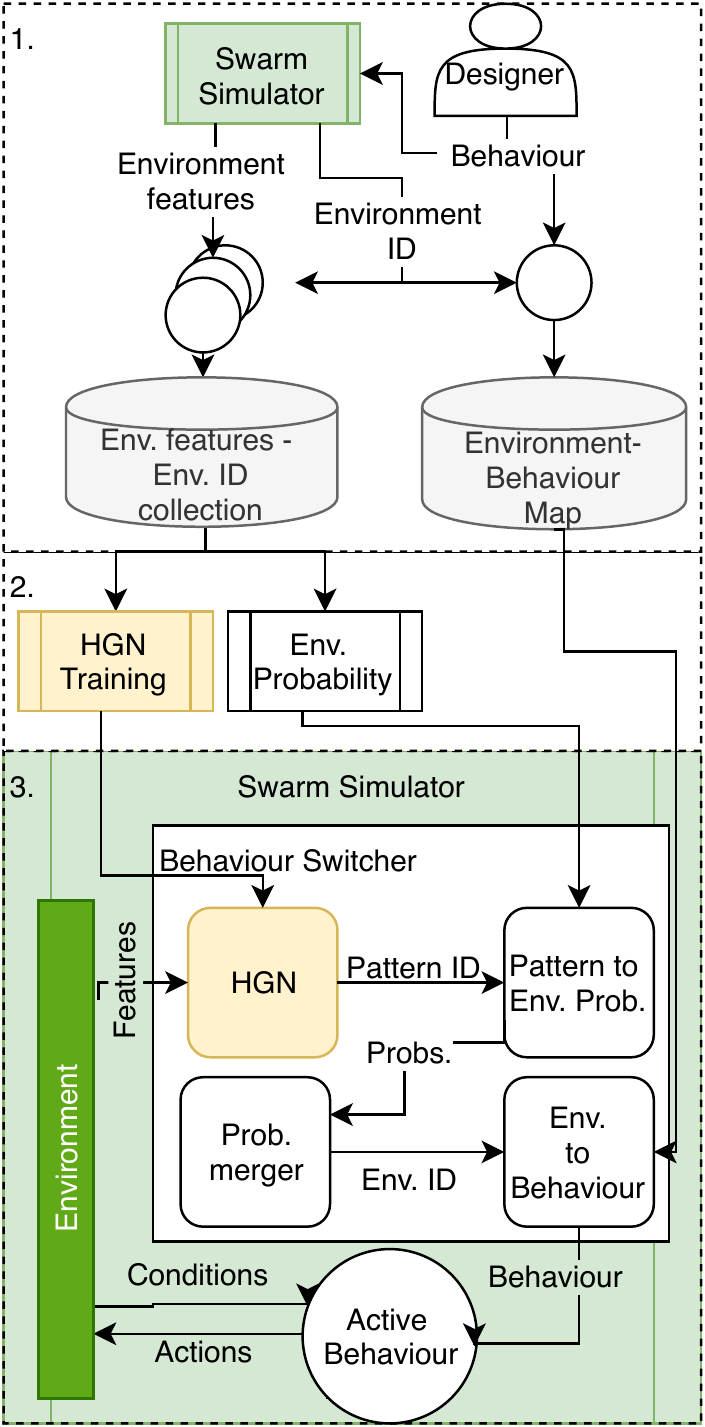}
	\caption{System Process: 1) Behaviour creation and feature extraction; 2) \gls{rhgn} and probability training; 3) swarm implementation with behaviour selection.
		Green denotes simulation environments. Yellow denotes \gls{hgn} processes. }
	\label{fig:processdiagram}
\end{figure}

In this paper, we equip each agent of a swarm with an \gls{rhgn} for the purpose of classifying environment state and switching the active behaviour to that listed as most appropriate. Figure~\ref{fig:processdiagram} depicts the proposed process of creating and training the \gls{rhgn} and how the agents utilise it for behaviour switching. This process is broken into three main sections: behaviour design, \gls{rhgn} training, and \gls{rhgn} swarm execution.

\subsubsection{Behaviour Creation}
In this first process, a number of behaviours are developed by a human designer and stored in the swarm agent repertoire. Each agent behaviour is developed such that when implemented in the full swarm an overall \textit{emergent} behaviour is achieved. These behaviours are developed in this study via a sequence of conditional executions, combined with the dynamic neighbourhood targeting system, discussed in \cite{smith2018data}. However, the \gls{rhgn} system is not restricted to such a behaviour mechanism, and any robotic control may be implemented. The behaviours developed explicitly for this study are further defined in Section~\ref{behaviours}.

After development, each behaviour is implemented in a number of environments.  The agents' observations during operation are extracted for later environment matching. These observations are made by each agent, in each time-step. Furthermore, the completed pattern database consists of observations from every behaviour in each example environment. Supplying the \gls{hgn} with such a volume of observations improves the pattern matching range and reoccurring observations allows greater accuracy in environment probability prediction.

\subsubsection{\gls{rhgn} Training}
When a significant behaviour repertoire has been created for the swarm, the \gls{rhgn} is trained with the aforementioned observations. During this training, the dynamic \gls{gn} rows are populated and the probability tuples are created and linked to \gls{rhgn} outputs, as discussed in Section~\ref{dynamciGN}. 

\subsubsection{\gls{rhgn} Implementation}
During operation, each swarm agent is supplied all behaviours, though  only has one active at any given time-step. The active behaviour controls the agent, as discussed in the behaviour creation phase, with incoming environment data determining agent actions. 

In addition to action control, environment data is fed to the \gls{rhgn} each time-step as a linear pattern. The \gls{hgn} output value is matched to the associated environment probability tuple. Each prediction tuple is stored in a collection, and after $T$ time-steps the predictions are averaged for behaviour (re)selection. This fusion prevents behaviour thrashing and improves environment match accuracy by predicting over a time series, not a single snapshot of the environment which may be limited or misleading due to sensing range and the stochastic environment. 
Furthermore, swarm agents share environment prediction with one-another while in communication range, similar to the swarm belief propagation of \cite{trianni2016emergence, Reina2015} and \cite{smith2018data}. These external prediction sets are added to the agents' collections and are equally incorporated into the fusion process. 

After all prediction fusion, a single environment is matched by each agent via the highest probability. The corresponding behaviour for the selected environment is implemented for $T$ time-steps, until the next environment match. It may be observed that as each agent makes an independent environment match, the swarm becomes behaviour-heterogeneous. Such a feature, combined with the prediction sharing, allows the swarm to accurately diversify temporally and spatially. That is, a portion of the swarm may selection one behaviour to overcome a local challenge, and then select another behaviour when this local challenge changes. Meanwhile, a separated portion of the swarm may utilise a third behaviour to overcome a different challenge.

\section{Experiment Design}
This section presents: the networking data-transfer task of the swarm; the \gls{rhgn} setting for this study; the three behaviours provided to the \gls{rhgn}, which are referred to as \gls{mb}; the two experiments of this study, \gls{rhgn} in manually designed environments and in randomly generated environments; and finally the result representation and analytical tools used to validate \gls{rhgn}. 

To validate the performance of \gls{rhgn} in the two experiments of this study, a random behaviour selector is additionally implemented for comparison. This algorithm, denoted as Rand(\gls{mb}), randomly selects an \gls{mb} for the entire swarm during start-up and does not change this behaviour for the full operation duration. This comparison between \gls{rhgn} and Rand(\gls{mb}) is conducted between resulting swarm fitness and environment identification accuracy. 

\subsection{Networking Task}
The swarm is tasked with facilitating a simplified data-transfer in which data-packets are transferred between network-nodes. This transfer consists of 1,000 data-packets and the accompanying acknowledgement-packets. The task is to be completed in a hostile environment, with obstacles restricting agent mobility and communication, and jamming devices heavily restricting communication within an area. 
Each agent is capable of storing ten packets in a buffer but is limited to transmitting or receiving one packet per time-step. Additionally, agents may move up to $0.22ms^{-1}$, with a time-step being 0.1 seconds.  Agents are equipped with a simple LIDAR, for obstacle detection, and simulate \gls{tof} and signal triangulation to detect neighbouring devices (fellow swarm agents, network-nodes and jammers). The data-transfer process of these experiments use the Gaussian shadow \gls{ldpl} model  \citep{rappaport1996wireless} with a signal loss exponent of 2.5, a transmission strength of 12dBm and a Gaussian standard deviation of 3. This model estimates the signal power loss over distance, determining if a communication attempt is successful. For a more in-depth discussion of the swarm agent operations the reader is referred to \cite{smith2017adaptive} and \cite{smith2018swarm}.

The data-transfer task is terminated when a time-limit $T$ is reached, or all packets reach their destinations. As this networking task aims to achieve high data throughput, the swarm's fitness is measured via, 
\begin{equation}
\mathrm{fit}=\frac{p_s}{p} - \frac{T_s}{T}, \mathrm{fit} \in \mathbb{R}:(-1,1) \label{fitness_equations} \vspace{-5pt}
\end{equation}
where $p_s$ is the packets that reached the destination within $T$ and $T_s$ is the time for all $p$ packets to be transferred. The termination criteria of this task results in either $\frac{T_s}{T}=1$, causing a negative fitness, or $p_s/p=1$, causing a positive fitness. In this study, $T$ is set to 50,000 time-steps, allowing ample environment traversal time.

\subsection{\gls{rhgn} Structure}
For the above data-transfer application, the \gls{rhgn} of this study distributes a pattern of 48 components into three sub-\gls{rhgn}, as used in \gls{dhgn}. These \glspl{rhgn} each focus on an aspect of the observations, which are: network conditions, packet statuses and neighbourhood conditions. The input pattern of these three \glspl{rhgn} are supplied in the appendix. It can be noted that some of these inputs may be superfluous for this study, however, the \gls{hgn} pattern matching process allows for such redundancy.

\gls{rhgn} pattern matching is conducted each time-step, and probabilities are fused for behaviour selection every 500 time-steps. Intra-swarm belief sharing has agents broadcast every 10 time-steps. During belief fusion, agents combine locally collected probabilities and neighbour probabilities with equal weighting. After behaviour selection, an agent's probability collection is cleared.

\subsection{Designed Behaviours} \label{behaviours}
The experiments of this study have the \gls{rhgn} and Rand(MB) agents equipped with a behaviour repertoire of three \glspl{mb}. Each of these behaviours has noted strengths and weaknesses and allows the swarm to operated in different conditions.

The first \gls{mb} evenly spaces agents between packet destination, packet origin, and other swarm agents. When in communication range, agents send packets to neighbours closer to said packet's destination.  Obstacles are avoided via a repulsive virtual force \citep{chang2003collision}, provided it does not prevent even distribution of agent spacing and no jamming avoidance is implemented. 

The second \gls{mb} focuses on wall circumvention via a ferrying behaviour \citep{zhao2004}. Agents travel towards other swarm members or packet destinations without the even spacing of \gls{mb}-1. Upon obstacle detection, the movement profile is dominated by repulsive and orbital virtual forces \citep{rezaee2011,chang2003collision} relative to the obstacle, as prior seen in \cite{smith2018swarm}. This motion profile continues until a closer obstacle is detected, the target is in communication range, or the agents are unable to move without collision. In the latter case, the agents reverse orbiting direction. This behaviour allows effective circumvention of signal blocking obstacles, but may overly manoeuvre around minor, avoidable obstacles.

The third \gls{mb} focuses on jammer avoidance. As with the first \gls{mb}, agents attempt even spatial distribution. However, if jamming noise is detected, agents move away from the noise source and avoid the area until 500 time-steps with no noise detection pass. This behaviour effectively overcomes jamming devices, however, false-positive jammer identification (which may occur when detecting other communication) causes unnecessary agent re-positioning and area avoidance. 

These three \gls{mb} are validated in six manually designed environments, Env. 1.1, 1.2, 1.3, 2.1, 2.2 and 2.3. This validation has each \gls{mb} operate in each environment and report the swarm fitness. The results are used to create the environment-behaviour mapping, shown in Table~\ref{staticbehaviours}, and the pattern-environment labels for \gls{rhgn} training.

\begin{table}
	\caption{Environment-Behaviour associations for \gls{rhgn}. }\label{staticbehaviours}
	\begin{tabular}{|l|l|l|l|l|l|l|}
		\hline    \textbf{Environment}  & 1.1 &1.2 & 1.3 &2.1 &2.2 &2.3 \\ \hline
		\textbf{Behaviour} &1&2&1&1&1&3 \\ \hline
	\end{tabular}
\end{table}

\subsection{Experiment one: Designed Environments}
For the manually designed environments, swarms of eight agents are implemented in eight environments with the three \gls{mb}, with the trained \gls{rhgn} and with Rand(\gls{mb}). These environments, designed to challenge the \gls{mb}, consist of the six listed in Table~\ref{staticbehaviours} and two additional environments, Env. 3.1 and 3.2, which the \gls{rhgn} has not been explicitly trained for. 

\begin{figure}
	\centering
	\subfigure[Env. 1.1]{\includegraphics[width=0.45\linewidth]{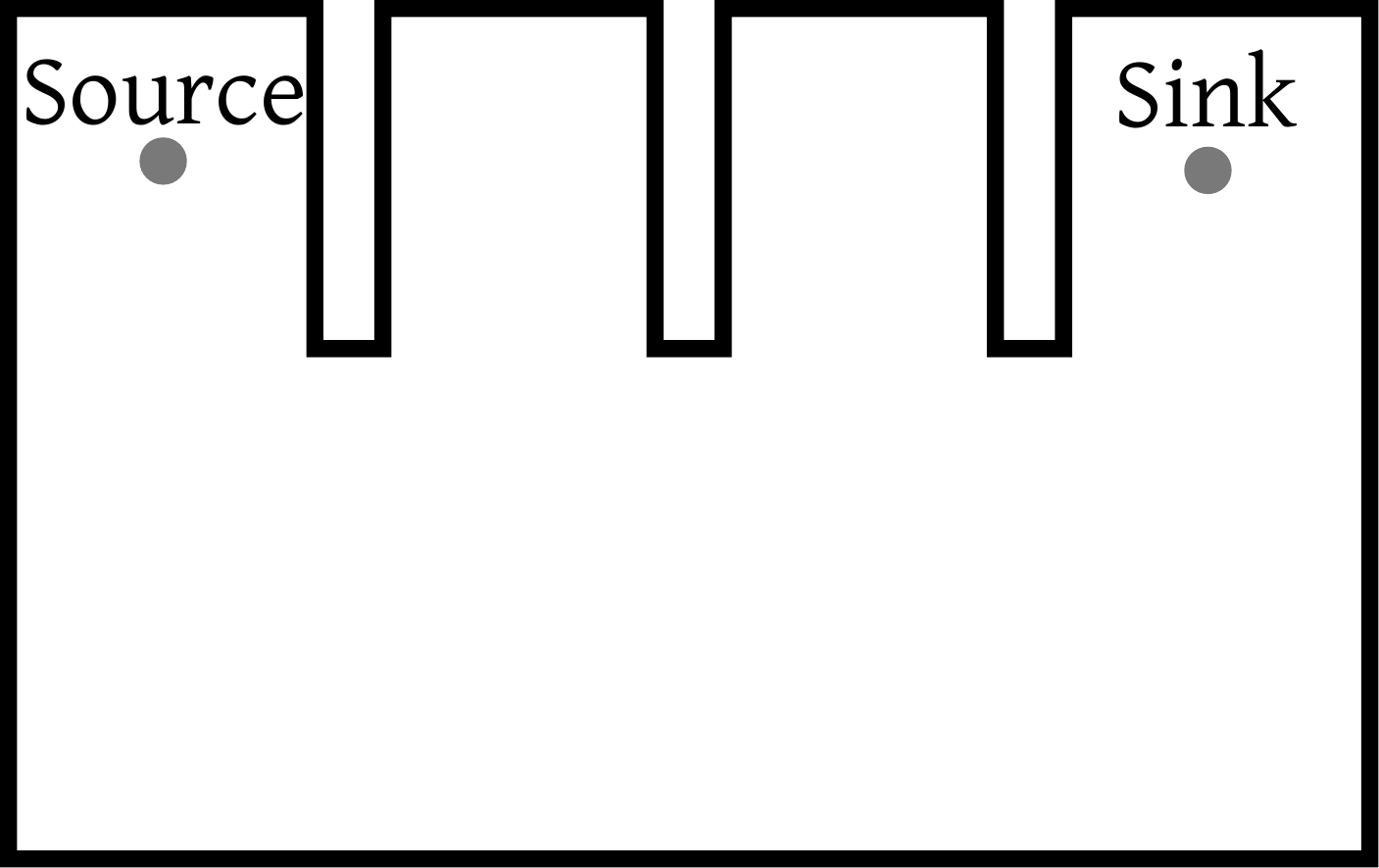}}    
	\subfigure[Env. 1.2]{\includegraphics[width=0.45\linewidth]{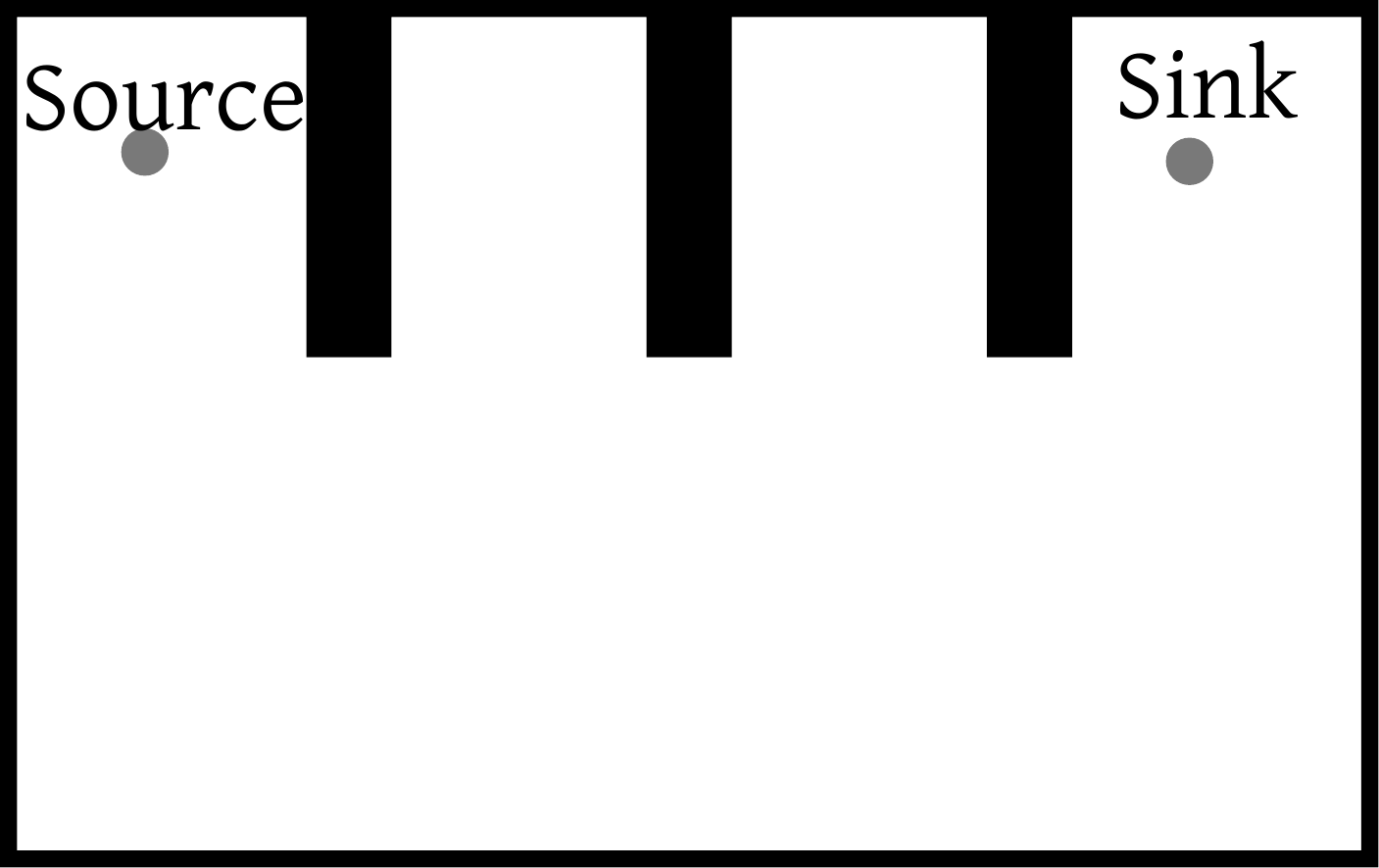}}    
	\subfigure[Env. 1.3]{\includegraphics[width=0.4\linewidth]{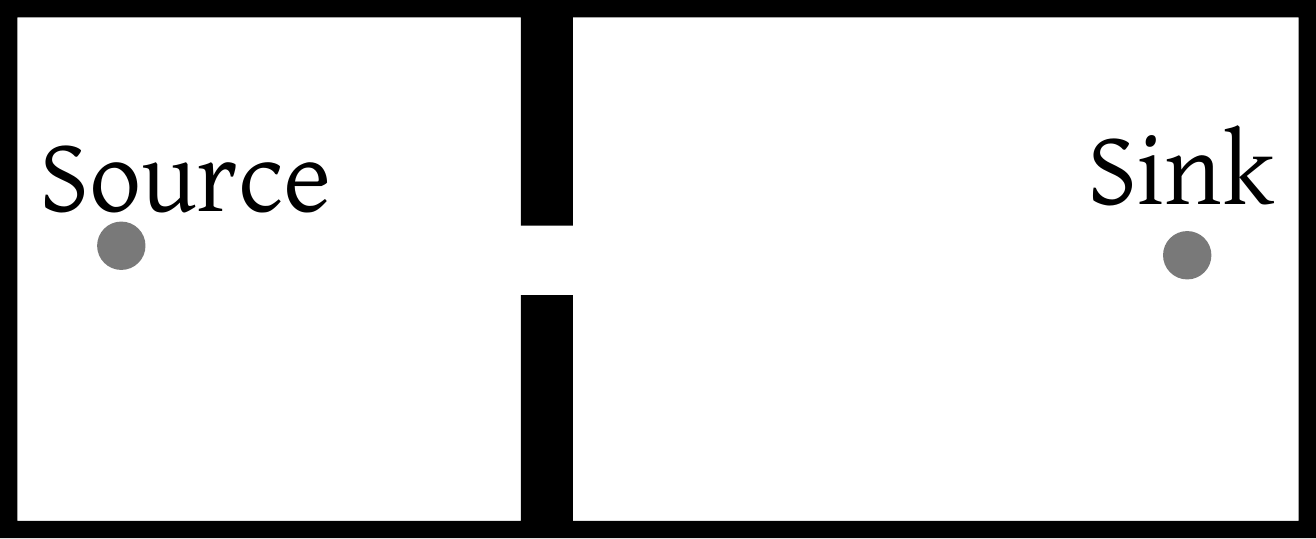}}
	\caption{Three obstacle configurations for the swarm to navigate. Lines are walls, with attenuation represented by wall thickness, grey circles are source and sink as labelled. }
	\label{fig:env1}
\end{figure}

The first three environments (1.1, 1.2 and 1.3), shown in Figure~\ref{fig:env1}, focus on obstacle configuration. In these operations, the swarm is tasked with mono-directional data-transfer between two network-nodes, labelled \textit{source} and \textit{sink}.  Env. 1.1 has the source and sink 60m apart, though separated by several thin walls with an attenuation factor of $5dBm$ each. The swarm agents may communicate through these walls, thus best performance is seen by \gls{mb}-1 positions the swarm directly between the source and sink. In contrast, Env. 1.2 has thicker walls which cause signal attenuation of $100dBm$, preventing data-transfer through them. As such, only \gls{mb}-2 can complete this operation by forming an arc around the walls. Finally, Env. 1.3 has the source and swarm start in a small walled area. To connect to the sink, the agents must pass through a small opening. This opening is sized so the wall avoiding actions of \gls{mb}-2 will be triggered and the swarm cannot pass. Therefore, the best performance is seen with any \gls{mb} other than \gls{mb}-2.

In relation to \gls{rhgn} challenges, these three environments present similar local obstacle information to each agent. The \gls{rhgn} must, therefore, recognise the neighbourhood and network conditions to identify the environment. Furthermore, as all agents may not be positioned such that the distinguishable network characteristics are observable, the intra-swarm belief propagation must be utilised for all agents to have correct environment identification.

Env. 2.1, 2.2 and 2.3 each have three network-nodes, \textit{A, B} and \textit{C} in an equilateral triangle with edges of 100m. Additionally, a jamming device is placed in the centre of this triangle. Topologically, these environments are identical, however, the distribution of packet origins and destinations, along with the jammer being active, are unique for each environment. These configurations are listed in Table~\ref{netwrokConfigs}. Env. 2.1 and 2.2 are seen to be similar settings, both seeing best performance by the non-jamming behaviours (\gls{mb}-1,2). Identification between these two environments is limited to packet distribution and is thus seen as a challenge for the \gls{rhgn} pattern matching. Env. 2.3 has the jammer active and thus requires the anti-jamming abilities of \gls{mb}-3. 
	\begin{table}
	\vspace*{-3mm}
	\caption{Network configuration of devices A, B and C for environment 2.1,2.2 and 2.3.}\label{netwrokConfigs}
	\begin{tabular}{|L{4.5mm}|| L{5mm}|L{4.5mm} || L{4.5mm}|L{4.5mm} || L{4.5mm}|L{4.5mm} || L{3mm}|}
		\hline
		\multirow{2}{*}{{\textbf{Env.}}} &
		\multicolumn{2}{c||}{\textbf{A}} &
		\multicolumn{2}{c||}{\textbf{B}} &
		\multicolumn{2}{c||}{\textbf{C}}&  \multirow{2}{*}{\rotatebox[origin=c]{90}{\parbox[c]{11mm}{ \textbf{Jammer}} }}\\ \cline{2-7}
		& \textit{Out (\%)} & \textit{In\ (\%)} & \textit{Out} (\%) & \textit{In\ (\%)} & \textit{Out (\%)} & \textit{In\ (\%)} &\\ \hline
		2.1& \footnotesize{100} &0 &0 &50 &0   &50    &$\times$ \\ \hline
		2.2& 33 & 33 &33 &33 &33 &33 &$\times$ \\ \hline
		2.3& 33 & 33 &33 &33 &33 &33 &$\checkmark$ \\ \hline
	\end{tabular}

\end{table}
\begin{figure}
	\vspace*{-3mm}
	\centering
	\includegraphics[width=0.4\linewidth]{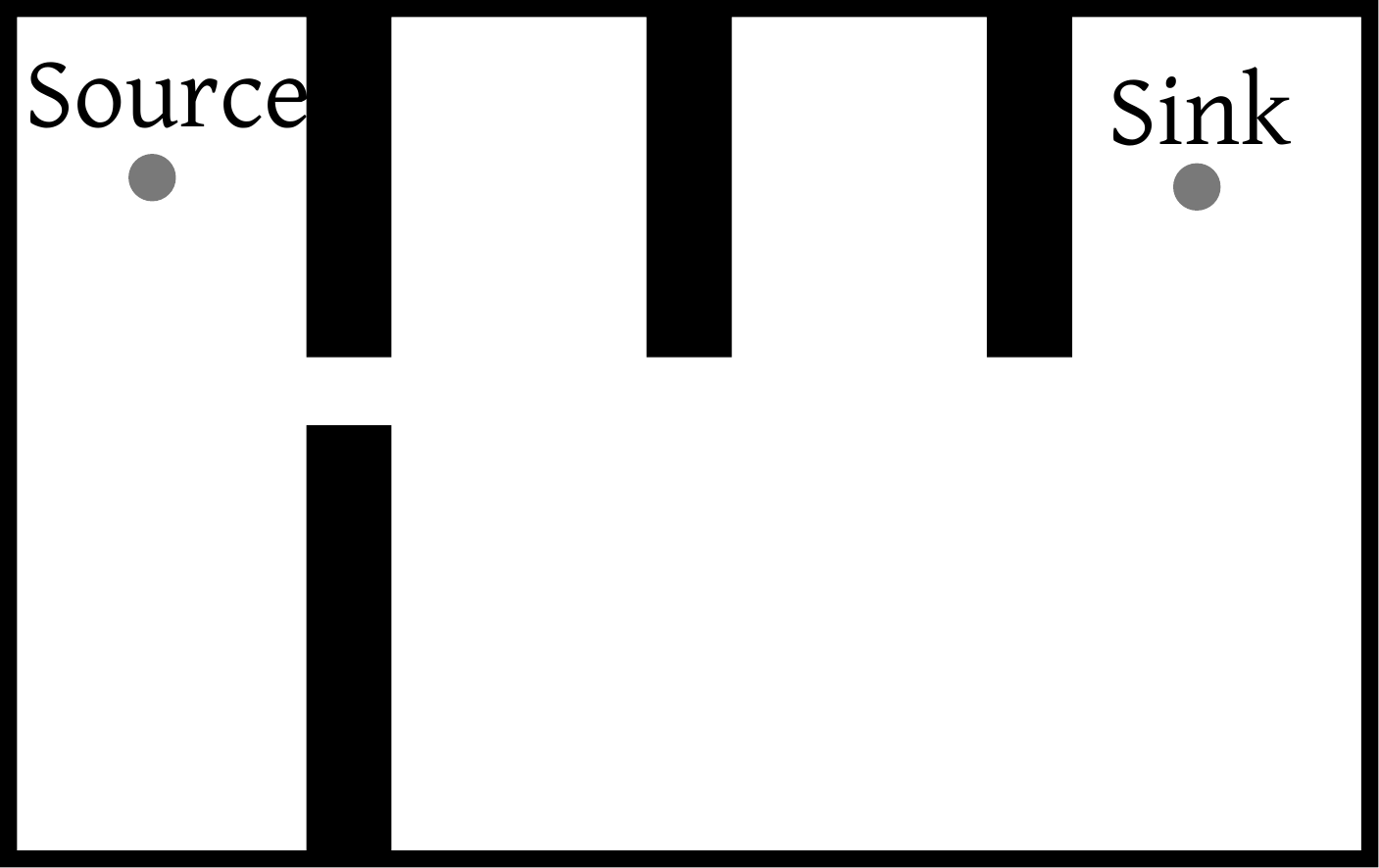}
	\caption{Env. 3.1: a combination of Env. 1.2 and Env. 1.3.}
	\label{fig:env13}
\end{figure}
Finally, for the additional Env. which \gls{rhgn} is not trained for, components of the prior Env. are combined to make spacial hybridisations.  Env. 3.1 combines the network restricting walls of Env. 1.2 and the narrow passage of Env. 1.3, as shown in Figure~\ref{fig:env13}. This environment requires agents to use \gls{mb}-1/\gls{mb}-3 to pass through the passage and \gls{mb}-2 to circumvent the walls. 
Similarly, Env. 3.2 combines 2.2 and 2.3. The packet distributed for 3.2 has all agents send to one another, however, the jammer is relocated from the centroid of A, B and C, to the centroid of A and B. Thus \gls{mb}-1 or -2 are optimal for connections A-C and C-B, and \gls{mb}-3 is required for connecting A-B.

To validate the abilities of \gls{rhgn} in these environments, a comparative study is conducted between the \gls{mb}, \gls{rhgn} and Rand(\gls{mb}) by implementing each swarm in the eight designed environments 50 times. Each implementation instance has a unique network randomisation seed and swarm member starting locations. Comparisons in the former six environments validate the environment matching ability of the \gls{rhgn}. Comparisons in the latter two environments explore \gls{rhgn} allowing spatial induced behaviour heterogeneity in the swarm.

\subsection{Experiment two: Generated Environments}

The second experiment of this study examines the ability of \gls{rhgn} to utilise the behaviour repertoire in prior unseen conditions.  This emulates the swarm having been training with controlled examples and then being deployed for real-world operation.

This experiment implements the \gls{mb}, \gls{rhgn} and Rand(MB) swarms in a further 50 environments which are randomly created. These environments are more challenging than the designed cases as the number of network-nodes is selected randomly from the set \{2,3,4\}, these nodes are mobile, and the data-transfer requirements between these nodes change during the operation. These requirement changes are discretised into data-transfer sub-operations and referred to as a \textit{network stage}. As an example of these stages, consider a network with two network-nodes, \textit{A} and \textit{B}, which undergoes two network stages. In the first stage, packets are sent from \textit{A} to \textit{B}. The swarm must make a mono-directional connection.  After these packets are transferred, the network requirement changes to stage two which has \textit{A} and \textit{B}  sending packets to each other. The swarm must reconfigure to facilitate this bi-directional connection. 
These network stages, and the network-node movement, requires the swarm to utilise multiple behaviours to facilitate all network conditions.

For the mobile network-nodes, waypoints are defined for each networking stage during environment generation. Waypoints are traversed by the mobile nodes at a speed of $0.11ms^{\neg 1}$, with the path between waypoints being a straight line. Upon reaching the final waypoint of the stage, the node becomes stationary.

For the network requirements of a stage, each network-node is assigned a percentage of the packets to be sent in the stage, and a percentage of these packets which should end at the node. During the creation of these values, it is enforced that all nodes are either a packet origin or destination in at least one stage. That is, no network-node is idle for the full operation.  

In this study, the number of network stages is limited to a random value from the set \{1,2,3\}. Each generated environment has the swarm transfer 1,000 packets, as in the prior experiment and these packets are evenly divided between the network stages.

In addition to the network-nodes, each environment generation also creates numerous obstacles and a jammer. Each obstacle is semi-randomly positioned in the $100m\times100m$ environment. This positioning is limited so the network-nodes' motions are not blocked. Similarly, the jammer is placed in the centre of the network-nodes but with the restriction of all network-nodes being outside jamming range. Thus the jammer is located in an area which the swarm agents are expected to enter, and thus will impact the swarm, but will not impact the network-nodes. Finally, the environment generator assigns an active or inactive state to the jammer for each network stage.

\subsection{Result representation}

For both environment experiments the \gls{rhgn} pattern matching is validated by comparing the fitness of the three \gls{mb} swarms, the \gls{rhgn} swarm and the Rand(\gls{mb}) swarm. This comparison presents the median and quartiles of each swarm fitness, and performs a Mann-Whitteny U-test between the \gls{rhgn} fitness results and the results of all other swarms. Additionally, this analysis examines the percentage of implementations which the \gls{rhgn} and Rand(\gls{mb}) achieve a fitnesses of at least 95\% of each \gls{mb}, and 95\% of the optimal \gls{mb} of that instance. This leniency of 5\% is given for \gls{rhgn}  as environment matching cannot be expected in the initial time-steps, due to agents not having adequate environment interaction and thus limited pattern based belief. This initial learning time leads to a small reduction in swarm fitness.

In addition to this fitness comparison, the accuracy of \gls{rhgn} environment prediction is explored for the first six designed environments. The accuracy is only tested in these cases as correct values are known and are not dependent on agent location. This analysis presents the error-rate over simulation time for each environment, averaged over the eight swarm agents and 50 environment instances. Additionally,  the one-versus-all accuracies and $F_1$ scores for each environment prediction are presented. These latter statistics are found across the eight swarm agents in all 300 swarm implementations (50 implementations of 6 environments). The accuracy, $Acc_{e}$, and $F_{1,e}$ scores of each environment, \textit{e}, are respectively  measured via, 
\begin{eqnarray}
Acc_e=\frac{TP_e+TN_e}{TP_e+TN_e+FP_e+FN_e}\\
F_{1,e}=\frac{2 \cdot TP_e}{2\cdot TP_e+ FP_e+FN_e}
\end{eqnarray}
where $TP_e$ is true positive selection of environment \textit{e}, $TN_e$ is true negative selection, and $FP_e$ and $FN_e$ are the respective false counterparts.

To analyse the spatial heterogeneity in Env. 3.1, 3.2, and the random environments, the environment maps are presented with markings denoting agent locations during behaviour selection and behaviour chosen. Each figure is the temporal concatenation of all time-steps over the operation or stage. For Env. 3.1 and 3.2, these figures are also concatenations of all 50 implementations. 

\section{Results and Discussion}
\subsection{Designed Environments}
\begin{figure}
	\centering
	\includegraphics[width=\linewidth]{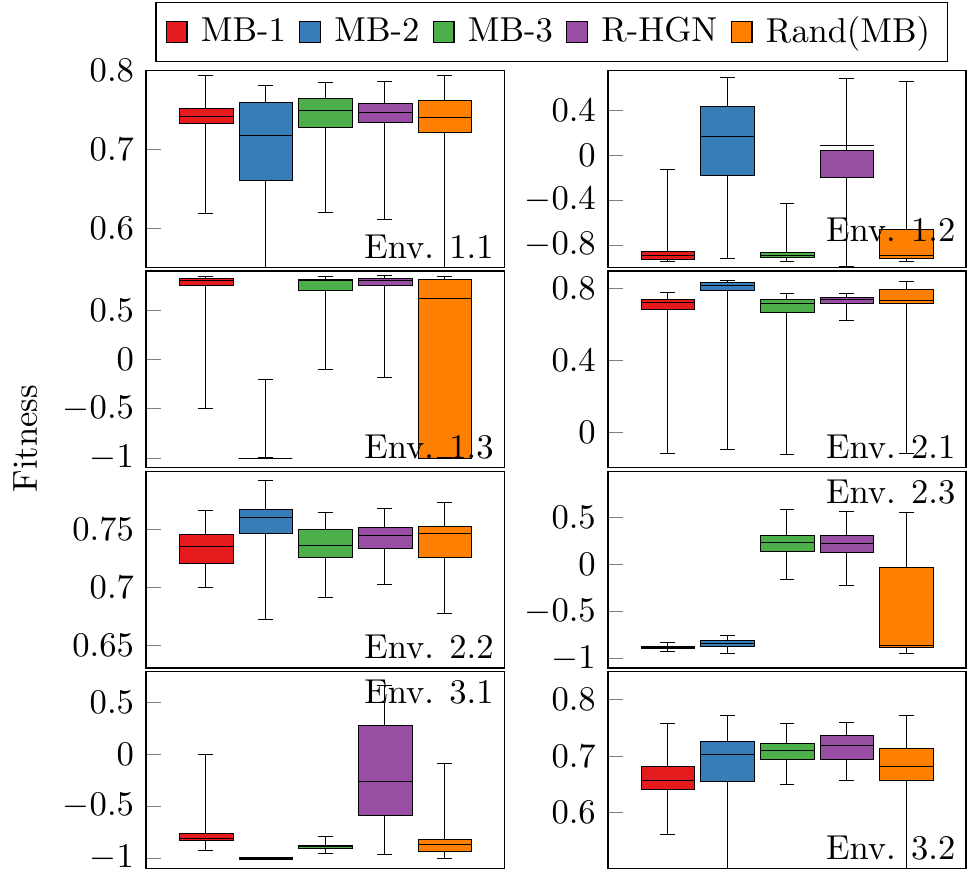}
	\caption{Box plot of swarm fitnesses using \glspl{mb}, \gls{rhgn} and Rand(MB) swarms in designed environment with direct training. \gls{rhgn} seen to achieve near equivalent performance to the optimal \gls{mb} in each environment.}
	\label{fig:test1}
\end{figure}

To begin the exploration of \gls{rhgn}, Figure~\ref{fig:test1} shows the performance of the three \gls{mb}, the trained \gls{rhgn} and Rand(MB) in the eight designed environments. In the six environments  which \gls{rhgn} was explicitly trained for (Env. 1.1-2.3), minimal performance difference between \gls{rhgn} and the optimal \gls{mb} is seen. Additionally, in three of these environments \gls{rhgn} shows significant improvement over Rand(MB).  In the two unseen environments (Env. 3.1 and 3.2), \gls{rhgn} outperforms all \gls{mb} and Rand(MB). These performance matching achievements by \gls{rhgn} are further supported in Table~\ref{tab:atLeast95}, showing in all environments \gls{rhgn} closely matches or outperforms the optimal \gls{mb} in at least 52\% of implementations, and in some environments this match rate is 100\%.  

	\begin{table}
	\centering
	\caption{Mann-Whitney U-Test between \glspl{mb} and \gls{rhgn}, and between Rand(MB) and \gls{rhgn}, in designed environment. In all but 2.1  \gls{rhgn} and the optimal \glspl{mb} have significant overlap. This shows the \gls{rhgn} is not only selecting, but correctly utilising the behaviours.} \label{tab:designedMan}
	\begin{tabular}{|l|c|c|c|c|}
		\hline
		&\textbf{\gls{mb}-1} & \textbf{\gls{mb}-2} & \textbf{\gls{mb}-3} & \textbf{Rand(MB)}\\ \hline
		\textbf{1.1}& 0.12                &\small$\ll\!\!0.01$ &0.84                &0.24 \\ \hline
		\textbf{1.2}& \small$\ll\!\!0.01$&0.88                 &\small$\ll\!\!0.01$ &\small$\ll\!\!0.01$    \\ \hline            
		\textbf{1.3}& 0.92                &\small$\ll\!\!0.01$ &0.22                &\small$\ll\!\!0.01$  \\ \hline
		\textbf{2.1}&\small$\ll\!\!0.01$ &\small$\ll\!\!0.01$ &\small$\ll\!\!0.01$ &0.51\\ \hline
		\textbf{2.2}&0.02                &\small$\ll\!\!0.01$ &0.12                &0.8\\ \hline
		\textbf{2.3}&\small$\ll\!\!0.01$ &\small$\ll\!\!0.01$ &0.65                &\small$\ll\!\!0.01$\\ \hline
		\textbf{3.1}&\small$\ll\!\!0.01$ &\small$\ll\!\!0.01$ &\small$\ll\!\!0.01$ &\small$\ll\!\!0.01$\\ \hline
		\textbf{3.2}&\small$\ll\!\!0.01$ &0.04                &0.13                &\small$\ll\!\!0.01$\\ \hline
	\end{tabular}
\end{table}
\begin{table}
	\centering
	\caption{Percentage of designed behaviour instance which \gls{rhgn} and Rnad(MB) achieves \mbox{$\ge 95\%$} of maximum \gls{mb} fitness.} \label{tab:atLeast95}
	\begin{tabular}{|c|c|c|c|c|c|c|}
		\hline
		&     \rotatebox[origin=c]{90}{\shortstack{\textbf{\gls{rhgn}:}\\\textbf{\gls{mb}-1}}} & 
		\rotatebox[origin=c]{90}{\shortstack{\textbf{\gls{rhgn}:}\\\textbf{\gls{mb}-2}}}&
		\rotatebox[origin=c]{90}{    \shortstack{\textbf{\gls{rhgn}:}\\\textbf{\gls{mb}-3}}} & \rotatebox[origin=c]{90}{\shortstack{\textbf{\gls{rhgn}:}\\ \textbf{Max(\gls{mb})}}} &  \rotatebox[origin=c]{90}{\shortstack{\textbf{Rand(\gls{mb})}:\\ \textbf{Max(\gls{mb})}}}\\ \hline
		\textbf{1.1}    &    98\%    &    100\%    &    98\%    &    98\%    &    88\%    \\ \hline
		\textbf{1.2}    &    96\%    &    52\%    &    96\%    &    52\%    &    28\%    \\ \hline
		\textbf{1.3}    &    86\%    &    100\%    &    82\%    &    80\%    &    46\%    \\ \hline
		\textbf{2.1}    &    98\%    &    68\%    &    100\%    &    66\%    &    62\%    \\ \hline
		\textbf{2.2}    &    100\%    &    100\%    &    100\%    &    100\%    &    98\%    \\ \hline
		\textbf{2.3}    &    100\%    &    60\%    &    60\%    &    60\%    &    28\%    \\ \hline
		\textbf{3.1}    &    80\%    &    100\%    &    92\%    &    80\%    &    40\%    \\ \hline
		\textbf{3.2}    &    100\%    &    100\%    &    96\%    &    100\%    &    86\%    \\ \hline
	\end{tabular}
\end{table}

For the behaviour matching in the designed environments which \gls{rhgn} is directly trained for, the most prominent success is seen in  Env. 1.2, 1.3 and 2.3. In these cases, some \gls{mb} are unable to complete the task and thus have median fitnesses below 0. In contrast, \gls{rhgn} utilised the optimal behaviour and achieves median fitnesses above 0 in all three cases. Furthermore, due to poor performing \gls{mb}, Rand(MB) sees considerably low median and quartile values and in Env. 1.3 sees considerable interquartile range . This shows \gls{rhgn} can effectively select the correct behaviour and overcome some \gls{mb} being invalid while Rand(MB) performance is significantly reduced by a poor \gls{mb} in the repertoire. Additionally, the superiority of \gls{rhgn} over the failing \gls{mb} and Rand(MB) is seen to be statistically significant in Table~\ref{tab:designedMan}, with low p-values shown for these results. Also, Table~\ref{tab:designedMan} shows high p-values for \gls{rhgn} and the optimal \gls{mb} in these cases. This supports \gls{rhgn}  correctly matching the environment to an \gls{mb} and achieving equivalent results.

In relation to \gls{rhgn} matching or outperforming the optimal \gls{mb} in Table~\ref{tab:atLeast95}, these results can be separated into three tiers of performance: high (Env. 1.1, 2.2 ), medium (Env. 1.3), and acceptable (Env. 1.2, 2.1, 2.3). This divide shows a relation between the performance of \gls{rhgn} in an environment and the percentage of repertoire \gls{mb} which are functional in that environment.  That is, high performance is seen when all three \gls{mb} are effective in the environment, medium performance is seen with $2/3$ being effective, and acceptable behaviour when the \gls{rhgn} is limited to one of the three behaviours to solve the task. This relationship is due to misinformed agents (agents observing patterns which are associated with other environments) utilising incorrect behaviours for the given environment and the impact such behaviour usage has on the swarm-wide operation. The high-performance cases can still utilise these incorrect behaviours, with the agents operating within expectation. The low performance cases see these behaviours negatively impact the swarm, with agents moving out of position for neighbour interaction or attempting actions known to fail in the environment.  That being said, all environments see \gls{rhgn} able to match or outperform the \glspl{mb} in at least half the implementations and \gls{rhgn} percentage are higher than Rand(MB) percentages in all environments. In the worst performance match for both behaviour selectors, Env. 1.2, \gls{rhgn} achieving 52\% which is 24\% higher than Rand(MB). Additionally, the overall match rate to the optimal \gls{mb} in these six Env. is 79.5\% for \gls{rhgn} and only  59.5\% for Rand(MB). This shows that \gls{rhgn} has achieved relatively accurate performance.

\begin{figure}
	\centering
	\subfigure[] {\includegraphics[width=0.4\linewidth]{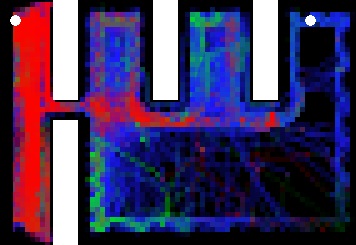}} \hspace*{0.01\linewidth}
	\subfigure[] {\includegraphics[width=0.4\linewidth]{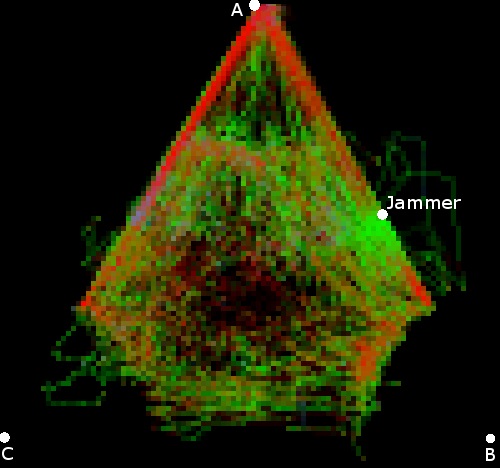}}
	\caption{Environment-Behaviour matching mapped over operation map for Env. 3.1 (a) and Env. 3.2 (b). Red selecting \gls{mb}-1; blue is \gls{mb}-2; green is \gls{mb}-3; white dots are network-node and jammers (as labelled), white squares are unreachable walled area.
		For Env. 3.1, strong use of \gls{mb}-1 around passage, \gls{mb}-2 around other  walls.
		For Env. 3.2, strong use of \gls{mb}-3 around jammer (between A and B), noticeable use of \gls{mb}-1 on A-C edge. Each figure is the combined mapping of all operation runs.}
	\label{fig:maps}
\end{figure}

For Env. 3.1, no \gls{mb} (and thus no Rand(MB)) in all 50 instances is seen to solve the data-transfer task within time-limit, $T$ . In contrast, \gls{rhgn} achieves 36\% operation success. Additionally, 80\% of operations seen \gls{rhgn} achieve at least 95\% of the optimal \gls{mb} performance in Table~\ref{tab:atLeast95} and of these 46 operations, \gls{rhgn} has an average fitness improvement of 860\% over the optimal \gls{mb}. These results are significantly higher than Rand(MB), with a match rate of only 40\%.  Furthermore, the higher fitness of \gls{rhgn} is seen to be statistically significant to all \gls{mb} and Rand(MB) in \Cref{tab:designedMan}. This higher performance due to \gls{rhgn} correctly matching the components of the environment is displayed in Figure~\ref{fig:maps} \textit{(a)} with the passageway to be identified as Env. 1.3, thus \gls{mb}-1 (red) is activated, and the wall sections are identified as Env. 1.2, causing \gls{mb}-2 (blue) to be used.

For Env. 3.2, both the median and upper-quartile  of \gls{rhgn} surpassed all \gls{mb} in Figure~\ref{fig:test1}, though this improvement is less significant than in Env. 3.1 and \Cref{tab:designedMan} shows this difference to be not statistically significant. On the other hand, Table~\ref{tab:atLeast95} shows 100\% of operations have \gls{rhgn}  reach at least 95\% of the optimal \gls{mb} fitness. In Figure~\ref{fig:maps} \textit{(b)} this higher performance is seen to be due to agents primarily utilising \gls{mb}-3 when close to the jammer, and primarily utilising \gls{mb}-1 when the jammer is sufficiently distant, as was predicted to be the case. Thus the \gls{rhgn} is correctly achieving a spatially heterogeneous behaviour selection. In relation to Rand(MB), the low performance of \gls{mb}-1 and \gls{mb}-2 are again reducing the median performance, and resulting in several failing instances. In contrast \gls{rhgn} has not instances which the swarm cannot transfer all data within $T$.

\begin{figure}
	\centering
	\includegraphics[width=1.0\linewidth]{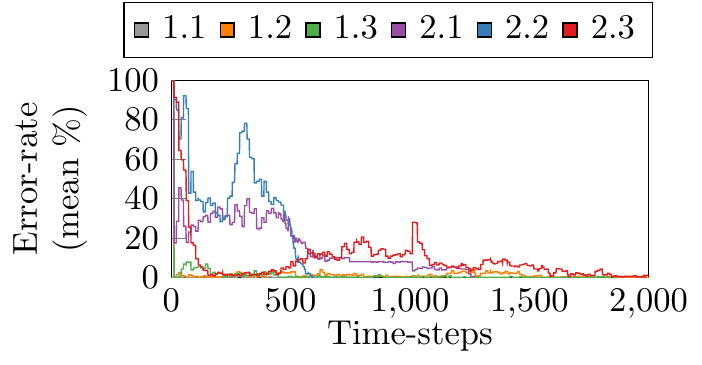}
	\caption{\gls{rhgn} error-rate (\%) of each designed Env. over time-steps. Error-rates/time-step vectors recorded for all 8 agent in 50 implementations. Presented values are mean of these 400 vectors. }
	\label{fig:designedenv-errorrate}
\end{figure}

In addition to exploring the swarm fitness, Figure~\ref{fig:designedenv-errorrate} shows the underlying \gls{hgn} environment matching mean error-rate over swarm operations for Env. 1.1-2.3. From this graph, it can be seen that most environments have some significant error-rates during early operation. However, as the agents distribute about the environment, and thus collect more informative pattern observations, these error-rates quickly decline. This is especially true with Env. 1.1-1.3, with error-rates dropping to $<5\%$ within $\sim10-100$ time-steps. For Env. 2.1-2.3, some errors continue throughout the operation. However, these error-rates are acceptably small, given how similar the environments appear from the agents' local observations; 2.2 is only distinguishable from 2.1 within the agents' local observations by a changing packet source; 2.3 is mistaken for 2.2 when the agents are not impacted by the jamming device. 

In addition to the error-rate over time, Table~\ref{tab:F_1_des} shows the environment matching accuracy of the swarm is between 92.4\% and 99.1\%  and the $F_1$ scores reach a top of 97.15\% and a bottom score of 57.81\%. These values show that \gls{rhgn} has a considerably high environment match rate over all 300 implementations. Furthermore, these values are considerably higher than the random environment matching algorithm, which is predicted to achieve a \{\textit{TP, FP, TN, FN} \} tuple of \{$1/36,5/36,25/36,5/36$\}, giving an accuracy of $72\%$ and  $F_1$ scores of  $16.7\%$ for all environments.

\begin{table*}
	\centering
	\caption{$F_1$ scores for the designed environments which \gls{rhgn} is explicitly trained for.}\label{tab:F_1_des}
	\begin{tabular}{|c|c|c|c|c|c|c|c|}
		\hline
		\textbf{Env.}&\textbf{1.1} &\textbf{1.2} &\textbf{1.3} &\textbf{2.1} &\textbf{2.2} &\textbf{2.3}\\ \hline
		\textbf{F1 (\%)}&93.3& 97.15&95.7&65.88&57.81&88.12\\ \hline
		\textbf{Accuracy (\%)}& 98.5& 98.5&99.1&92.4&92.4&92.5 \\ \hline 
	\end{tabular}
	\vspace*{-5pt}
\end{table*}

To conclude this exploration of \gls{rhgn} in the designed environments, it is shown that \gls{rhgn} can aptly identify the environment, or partial-environment, and utilises the associated behaviour for optimal swarm operation. This correct identification is confirmed by high accuracy and $F_1$ scores.  \gls{rhgn} allowed the swarm to hold a median fitness above 0 for all environments, and a median fitnesses higher than any \gls{mb} in composite environments. This shows that although \gls{mb} are capable solutions to the data-transfer task in the intended solutions, they have limited flexibility and cannot be used for any implementation. \gls{rhgn} overcomes this flexibility limitation, allowing the swarm to achieve high fitness in a wider range of implementations, including more complex environments.

\subsection{Generated Environments}

\begin{figure}
	\centering
	\includegraphics[width=\linewidth]{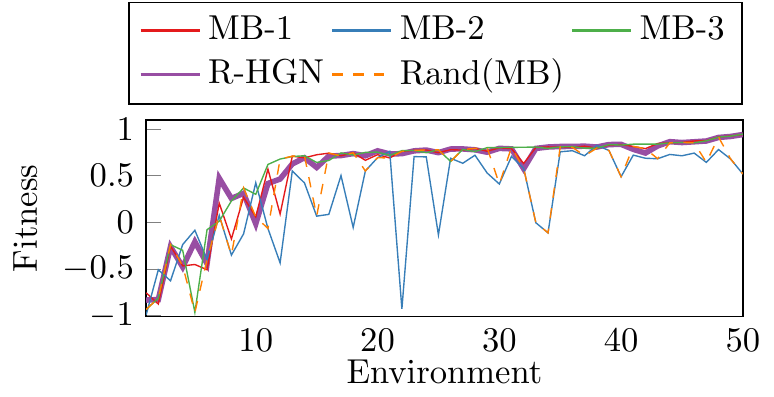}
	\caption{Fitness of three \gls{mb}, \gls{rhgn} and Rand(MB) in 50 generated environments. Environments ordered by maximum \gls{mb} fitness for clarity. \gls{rhgn} holds close fitness to optimal \gls{mb} in all environments.}
	\label{fig:genres}
\end{figure}

To further explore the ability of \gls{rhgn}, Figure~\ref{fig:genres} shows the fitnesses of the three \gls{mb}, \gls{rhgn} and Rand(MB) for 50 randomly generated environments, ordered by the optimal \gls{mb} fitness. Table~\ref{tab:genMan} presents the associated Mann-Whitney U-Test results. 
As can be seen, \gls{rhgn} continues to achieve swarm fitnesses similar to the optimal \gls{mb} in the majority of environments. Additionally, in several environments, the spatial and temporal behaviour diversification of \gls{rhgn} allows fitnesses greater than a single \gls{mb}. However, this fitness improvement is limited, and a notable correlation exists between \gls{rhgn} performances and the potential of the behaviour repertoire. That is, in all environments, \gls{rhgn} is only able to produce swarm fitnesses slightly higher than the best \gls{mb}. In relation to Rand(MB), this correlation is seen to exist between both the optimal and sub-optimal behaviours; several environments see Rand(MB) select an \gls{mb} with lower fitness than the other \gls{mb} which leading to considerably lower performance than \gls{rhgn}. In relation to statistical similarity,  Table~\ref{tab:genMan} show \gls{rhgn} again has low overlap with the failing \gls{mb}-2 and Rand(MB), but very high overlap with \gls{mb}-1 and \gls{mb}-3, which are optimal in most environments. 

	\begin{table}
	\centering
	\caption{Mann-Whitney U-Test  between \glspl{mb} and \gls{rhgn}, and between Rand(MB) and \gls{rhgn}, in generated environments. \gls{rhgn} has high statistical overlap with \gls{mb}-1 and \gls{mb}-3  and low overlap with \gls{mb}-2 and Rand(MB).} \label{tab:genMan}
	\begin{tabular}{|l|c|c|c|c|c|c|}
		\hline
		\textbf{\gls{mb}-1} & \textbf{\gls{mb}-2} & \textbf{\gls{mb}-3} & \textbf{Rand(MB)}\\ \hline
		0.91 & 0.01 & 0.66&0.05 \\ \hline
	\end{tabular}
\end{table}

This appropriate behaviour usage by \gls{rhgn} is further demonstrated in \Cref{tab:table_gen} with the percentage of environments with  $\ge 95\%$ \gls{mb} fitness again shown. These results show that \gls{rhgn} achieved close to optimal behaviour in 78\% of environments, a value significantly high given the underlying \gls{hgn} has no prior experience with the 50 generated environments. Additionally, as the optimal behaviour changes for each environment, \gls{rhgn} achieves significantly higher performance than each \gls{mb} used in isolation. This is seen from the fitness match rate being up to 92\%. Finally, \gls{rhgn} achieved far higher match rates than Rand(MB), which reaches only 56\%.  

\begin{figure}
	\centering
	\includegraphics[width=0.4\linewidth]{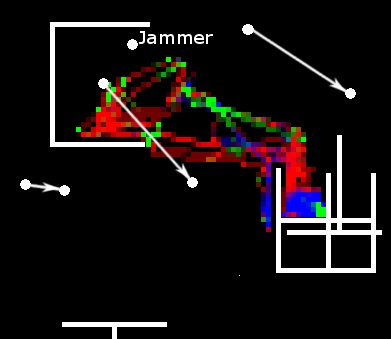}%
	\hspace*{1mm}        \includegraphics[width=0.4\linewidth]{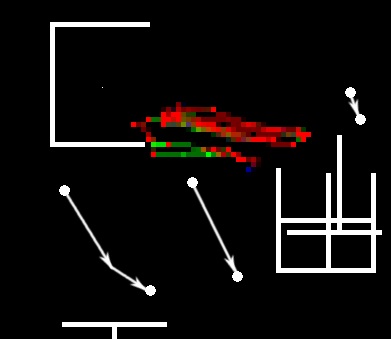}%
	\caption{Environment-Behaviour matching map over generated environment with two communication stages. Colouring as in Figure~\ref{fig:maps}, white arrows depict network-node motion with start and end locations. Environment matches  shown to allow both spacial and temporal behaviour heterogeneity to solve dynamic task. }
	\label{fig:gen1}
\end{figure}

Finally, to demonstrate the \gls{rhgn} performance in these challenging, generated environments,  Figure~\ref{fig:gen1} again maps the behaviours over the environment landscape for a noteworthy environment. This environment has two network stages and thus each is depicted separately. In stage 1, left, the jammer is active and the swarm is seen to identify this pattern component and primarily utilise \gls{mb}-3. Additionally, agents trapped by the obstacles in the lower-right area are switched to \gls{mb}-2, circumventing the obstacle. In stage 2, right, the jammer is disabled and the swarm is positioned away from obstacles. As such \gls{mb}-1 sees primary usage.
This diverse and dynamic environment shows \gls{rhgn} allows the swarm to heterogeneously utilise the behaviours as required across both time and space.

From this exploration, it is concluded that \gls{rhgn} can effectively guide the swarm agents' behaviour usage, even when the environments are unknown and distorted pattern matching is required. Additionally, \gls{rhgn} continues to function when the environments are more complex than training conditions.

	\begin{table}
	\centering
	\caption{Percentage of generated environments which \gls{rhgn} and Rand(MB) achieves $ \ge 95\%$ fitness of \glspl{mb}. All \gls{mb} and max(MB) shows strong matching performance of \gls{rhgn}}. \label{tab:table_gen}
	\begin{tabular}{|c|c|c|c|c|}
		\hline
		\rotatebox[origin=c]{90}{\shortstack{\textbf{\gls{rhgn}:}\\\textbf{\gls{mb}-1}}} & 
		\rotatebox[origin=c]{90}{\shortstack{\textbf{\gls{rhgn}:}\\\textbf{\gls{mb}-2}}}&
		\rotatebox[origin=c]{90}{    \shortstack{\textbf{\gls{rhgn}:}\\\textbf{\gls{mb}-3}}} & \rotatebox[origin=c]{90}{\shortstack{\textbf{\gls{rhgn}:}\\ \textbf{Max(\gls{mb})}}} &  \rotatebox[origin=c]{90}{\shortstack{\textbf{Rand(\gls{mb})}:\\ \textbf{Max(\gls{mb})}}}\\ \hline
		92\% & 90\% &84\% & 78\% &56\% \\ \hline    
	\end{tabular}
\end{table}

\ \\

Experiment data-sets have been made available at Monash FigShare, doi: \url{https://doi.org/10.26180/5d356d93a2a60}.

\section{Conclusion}
This paper presents an extension of \gls{hgn} for online behaviour selection in a robotic swarm tasked with data-transfer between networking devices. This \gls{hgn} extension, named \acrlong{rhgn}, allows pattern matching of real-value inputs without costly memory or computation consumption and outputs match probabilities for more effective temporal and intra-swarm prediction fusion.

Using the proposed \gls{rhgn} with three manually designed behaviours, the swarm was implemented in a number of human-designed and randomly generated environments. These \gls{rhgn} swarm performances were compared against the individual behaviours and a random behaviour selector.

In relation to the designed environments, it was found that the \gls{rhgn} closely matched the performance of the optimal behaviour when directly trained in the environment. 
 Additionally, \gls{rhgn} achieved an environment-matching accuracy up to 99.1\% and a top $F_1$ score of 97.15\%,  far higher than the random behaviour selector.
For environments which were a concatenation of the training conditions, \gls{rhgn} outperformed all individual behaviours and the random behaviour selector.

In relation to the generated environments, which introduced more challenging versions of the swarm operation, the \gls{rhgn} continued to match or outperform the supplied behaviours and the random behaviour selector. However, the \gls{rhgn} driven swarm performance was limited by the potential of the supplied behaviours. 

This limitation leads to our future work which will combine our prior studies in swarm behaviour creation \citep{smith2018swarm} with this study of behaviour selection. Such a combination aims to produce a swarm behaviour controller which may have the swarm select from the behaviour repertoire during operation, followed by autonomously create a new, more appropriate, behaviour for the environment post operation. Such a combination will allow continued repertoire extension as the swarm is deployed in more environments, allowing life-long behaviour learning.

	\section*{Acknowledgement}
	Funding for this research was provided by Cyber and Electronic Warfare Division, Defence Science and Technology Group, Commonwealth of Australia.
	
	\section*{Conflict of Interest}
	None.

\section*{Credit Authorship Contribution Statement}	
\textbf{Phillip Smith:} Software, Conducted Simulations, Interpreted Results, Writing - Original draft preparation.
 \textbf{Aldeida Aleti:} Project Supervisor, Writer - feedback. 
\textbf{Cheng-Siong Lee:} Project Supervisor, Writer - feedback.
\textbf{Robert Hunjet:} Project Supervisor, Writer - feedback, Conceptualization of this study.
\textbf{Asad Khan:} Project Supervisor, Writer - feedback,
Conceptualization of this study.

\bibliographystyle{cas-model2-names}

\bibliography{thebib}   

\begin{thebibliography}{30}
\expandafter\ifx\csname natexlab\endcsname\relax\def\natexlab#1{#1}\fi
\providecommand{\url}[1]{\texttt{#1}}
\providecommand{\href}[2]{#2}
\providecommand{\path}[1]{#1}
\providecommand{\DOIprefix}{doi:}
\providecommand{\ArXivprefix}{arXiv:}
\providecommand{\URLprefix}{URL: }
\providecommand{\Pubmedprefix}{pmid:}
\providecommand{\doi}[1]{\href{http://dx.doi.org/#1}{\path{#1}}}
\providecommand{\Pubmed}[1]{\href{pmid:#1}{\path{#1}}}
\providecommand{\bibinfo}[2]{#2}
\ifx\xfnm\relax \def\xfnm[#1]{\unskip,\space#1}\fi
\bibitem[{Baum and Haussler(1989)}]{baum1989size}
\bibinfo{author}{Baum, E.B.}, \bibinfo{author}{Haussler, D.},
  \bibinfo{year}{1989}.
\newblock \bibinfo{title}{What size net gives valid generalization?}, in:
  \bibinfo{booktitle}{Advances in neural information processing systems}, pp.
  \bibinfo{pages}{81--90}.
\bibitem[{Burke et~al.(2013)Burke, Gendreau, Hyde, Kendall, Ochoa, Ozcan and
  Qu}]{Burke2013}
\bibinfo{author}{Burke, E.K.}, \bibinfo{author}{Gendreau, M.},
  \bibinfo{author}{Hyde, M.}, \bibinfo{author}{Kendall, G.},
  \bibinfo{author}{Ochoa, G.}, \bibinfo{author}{Ozcan, E.},
  \bibinfo{author}{Qu, R.}, \bibinfo{year}{2013}.
\newblock \bibinfo{title}{Hyper-heuristics: a survey of the state of the art}.
\newblock \bibinfo{journal}{Journal of the Operational Research Society}
  \bibinfo{volume}{64}, \bibinfo{pages}{1695--1724}.
\newblock \DOIprefix\doi{10.1057/jors.2013.71}.
\bibitem[{Burke et~al.(2006)Burke, Petrovic and Qu}]{burke2006case}
\bibinfo{author}{Burke, E.K.}, \bibinfo{author}{Petrovic, S.},
  \bibinfo{author}{Qu, R.}, \bibinfo{year}{2006}.
\newblock \bibinfo{title}{Case-based heuristic selection for timetabling
  problems}.
\newblock \bibinfo{journal}{Journal of Scheduling} \bibinfo{volume}{9},
  \bibinfo{pages}{115--132}.
\bibitem[{Chang et~al.(2003)Chang, Shadden, Marsden and
  Olfati-Saber}]{chang2003collision}
\bibinfo{author}{Chang, D.E.}, \bibinfo{author}{Shadden, S.C.},
  \bibinfo{author}{Marsden, J.E.}, \bibinfo{author}{Olfati-Saber, R.},
  \bibinfo{year}{2003}.
\newblock \bibinfo{title}{Collision avoidance for multiple agent systems}, in:
  \bibinfo{booktitle}{2003 IEEE Conference on Decision and Control},
  \bibinfo{publisher}{IEEE}. pp. \bibinfo{pages}{539--543}.
\bibitem[{Glover and Kochenberger(2006)}]{Kochenberger2003}
\bibinfo{author}{Glover, F.W.}, \bibinfo{author}{Kochenberger, G.A.},
  \bibinfo{year}{2006}.
\newblock \bibinfo{title}{Handbook of metaheuristics}.
  volume~\bibinfo{volume}{57}.
\newblock \bibinfo{publisher}{Springer Science \& Business Media}.
\bibitem[{Hagenauer and Helbich(2017)}]{hagenauer2017comparative}
\bibinfo{author}{Hagenauer, J.}, \bibinfo{author}{Helbich, M.},
  \bibinfo{year}{2017}.
\newblock \bibinfo{title}{A comparative study of machine learning classifiers
  for modeling travel mode choice}.
\newblock \bibinfo{journal}{Expert Systems with Applications}
  \bibinfo{volume}{78}, \bibinfo{pages}{273--282}.
\bibitem[{Hettiarachchige et~al.(2018)Hettiarachchige, Khan and
  Barca}]{hettiarachchige2018multi}
\bibinfo{author}{Hettiarachchige, Y.}, \bibinfo{author}{Khan, A.},
  \bibinfo{author}{Barca, J.C.}, \bibinfo{year}{2018}.
\newblock \bibinfo{title}{Multi-object tracking of swarms with active target
  avoidance}, in: \bibinfo{booktitle}{2018 15th International Conference on
  Control, Automation, Robotics and Vision (ICARCV)},
  \bibinfo{organization}{IEEE}. pp. \bibinfo{pages}{1204--1209}.
\bibitem[{Khan and Ramachandran(2002)}]{khan2002peer}
\bibinfo{author}{Khan, A.}, \bibinfo{author}{Ramachandran, V.},
  \bibinfo{year}{2002}.
\newblock \bibinfo{title}{A peer-to-peer associative memory network for
  intelligent information systems}, in: \bibinfo{booktitle}{ACIS 2002
  Proceedings}, pp. \bibinfo{pages}{6--17}.
\bibitem[{Kim(2008)}]{kim2008comparison}
\bibinfo{author}{Kim, Y.S.}, \bibinfo{year}{2008}.
\newblock \bibinfo{title}{Comparison of the decision tree, artificial neural
  network, and linear regression methods based on the number and types of
  independent variables and sample size}.
\newblock \bibinfo{journal}{Expert Systems with Applications}
  \bibinfo{volume}{34}, \bibinfo{pages}{1227--1234}.
\bibitem[{Leng et~al.(2017)Leng, Yu, Zhang, Zhang, He and Zhou}]{leng2017task}
\bibinfo{author}{Leng, Y.}, \bibinfo{author}{Yu, C.}, \bibinfo{author}{Zhang,
  W.}, \bibinfo{author}{Zhang, Y.}, \bibinfo{author}{He, X.},
  \bibinfo{author}{Zhou, W.}, \bibinfo{year}{2017}.
\newblock \bibinfo{title}{Task-oriented hierarchical control architecture for
  swarm robotic system}.
\newblock \bibinfo{journal}{Natural Computing} \bibinfo{volume}{16},
  \bibinfo{pages}{579--596}.
\bibitem[{Mahmood et~al.(2008)Mahmood, Muhamad~Amin and Khan}]{dhgn}
\bibinfo{author}{Mahmood, R.}, \bibinfo{author}{Muhamad~Amin, A.H.},
  \bibinfo{author}{Khan, A.}, \bibinfo{year}{2008}.
\newblock \bibinfo{title}{A lightweight, fast and efficient distributed
  hierarchical graph ne uron-based pattern classifier}.
\newblock \bibinfo{journal}{International Journal of Intelligent Engineering
  and Systems} \bibinfo{volume}{1}, \bibinfo{pages}{9--17}.
\newblock \DOIprefix\doi{10.22266/ijies2008.1231.02}.
\bibitem[{Misir et~al.(2009)Misir, Wauters, Verbeeck and Berghe}]{misir2009}
\bibinfo{author}{Misir, M.}, \bibinfo{author}{Wauters, T.},
  \bibinfo{author}{Verbeeck, K.}, \bibinfo{author}{Berghe, G.V.},
  \bibinfo{year}{2009}.
\newblock \bibinfo{title}{A new learning hyper-heuristic for the traveling
  tournament problem}, in: \bibinfo{booktitle}{Proceedings of the 8th
  Metaheuristic International Conference (MIC09). Hamburg: Germany},
  \bibinfo{organization}{Citeseer}.
\bibitem[{Nagavalli et~al.(2017)Nagavalli, Chakraborty and
  Sycara}]{nagavalli2017automated}
\bibinfo{author}{Nagavalli, S.}, \bibinfo{author}{Chakraborty, N.},
  \bibinfo{author}{Sycara, K.}, \bibinfo{year}{2017}.
\newblock \bibinfo{title}{Automated sequencing of swarm behaviors for
  supervisory control of robotic swarms}, in: \bibinfo{booktitle}{2017 IEEE
  International Conference on Robotics and Automation (ICRA)},
  \bibinfo{organization}{IEEE}. pp. \bibinfo{pages}{2674--2681}.
\bibitem[{Nasution and Khan(2008)}]{nasution2008hierarchical}
\bibinfo{author}{Nasution, B.B.}, \bibinfo{author}{Khan, A.I.},
  \bibinfo{year}{2008}.
\newblock \bibinfo{title}{A hierarchical graph neuron scheme for real-time
  pattern recognition}.
\newblock \bibinfo{journal}{IEEE Transactions on Neural Networks}
  \bibinfo{volume}{19}, \bibinfo{pages}{212--229}.
\bibitem[{Pappa et~al.(2013)Pappa, Ochoa, Hyde, Freitas, Woodward and
  Swan}]{Pappa2014}
\bibinfo{author}{Pappa, G.L.}, \bibinfo{author}{Ochoa, G.},
  \bibinfo{author}{Hyde, M.R.}, \bibinfo{author}{Freitas, A.A.},
  \bibinfo{author}{Woodward, J.}, \bibinfo{author}{Swan, J.},
  \bibinfo{year}{2013}.
\newblock \bibinfo{title}{Contrasting meta-learning and hyper-heuristic
  research: the role of evolutionary algorithms}.
\newblock \bibinfo{journal}{Genetic Programming and Evolvable Machines}
  \bibinfo{volume}{15}, \bibinfo{pages}{3--35}.
\newblock \DOIprefix\doi{10.1007/s10710-013-9186-9}.
\bibitem[{Rappaport et~al.(1996)}]{rappaport1996wireless}
\bibinfo{author}{Rappaport, T.S.}, et~al., \bibinfo{year}{1996}.
\newblock \bibinfo{title}{Wireless communications: principles and practice}.
  volume~\bibinfo{volume}{2}.
\newblock \bibinfo{publisher}{prentice hall PTR New Jersey}.
\bibitem[{Reina et~al.(2015)Reina, Valentini, Fernández-Oto, Dorigo and
  Trianni}]{Reina2015}
\bibinfo{author}{Reina, A.}, \bibinfo{author}{Valentini, G.},
  \bibinfo{author}{Fernández-Oto, C.}, \bibinfo{author}{Dorigo, M.},
  \bibinfo{author}{Trianni, V.}, \bibinfo{year}{2015}.
\newblock \bibinfo{title}{A design pattern for decentralised decision making}.
\newblock \bibinfo{journal}{PLOS ONE} \bibinfo{volume}{10},
  \bibinfo{pages}{1--18}.
\newblock \URLprefix \url{https://doi.org/10.1371/journal.pone.0140950},
  \DOIprefix\doi{10.1371/journal.pone.0140950}.
\bibitem[{Rezaee and Abdollahi(2014)}]{rezaee2011}
\bibinfo{author}{Rezaee, H.}, \bibinfo{author}{Abdollahi, F.},
  \bibinfo{year}{2014}.
\newblock \bibinfo{title}{A decentralized cooperative control scheme with
  obstacle avoidance for a team of mobile robots}.
\newblock \bibinfo{journal}{IEEE Transactions on Industrial Electronics}
  \bibinfo{volume}{61}, \bibinfo{pages}{347--354}.
\bibitem[{Smallwood and Sondik(1973)}]{smallwood1973optimal}
\bibinfo{author}{Smallwood, R.D.}, \bibinfo{author}{Sondik, E.J.},
  \bibinfo{year}{1973}.
\newblock \bibinfo{title}{The optimal control of partially observable markov
  processes over a finite horizon}.
\newblock \bibinfo{journal}{Operations research} \bibinfo{volume}{21},
  \bibinfo{pages}{1071--1088}.
\bibitem[{Smith et~al.(2017)Smith, Hunjet, Aleti and Barca}]{smith2017adaptive}
\bibinfo{author}{Smith, P.}, \bibinfo{author}{Hunjet, R.},
  \bibinfo{author}{Aleti, A.}, \bibinfo{author}{Barca, J.C.},
  \bibinfo{year}{2017}.
\newblock \bibinfo{title}{Adaptive data transfer methods via policy evolution
  for uav swarms}, in: \bibinfo{booktitle}{2017 27th International
  Telecommunication Networks and Applications Conference (ITNAC)},
  \bibinfo{organization}{IEEE}. pp. \bibinfo{pages}{1--8}.
\bibitem[{Smith et~al.(2018a)Smith, Hunjet, Aleti, Barca
  et~al.}]{smith2018data}
\bibinfo{author}{Smith, P.}, \bibinfo{author}{Hunjet, R.},
  \bibinfo{author}{Aleti, A.}, \bibinfo{author}{Barca, J.C.}, et~al.,
  \bibinfo{year}{2018}a.
\newblock \bibinfo{title}{Data transfer via uav swarm behaviours: Rule
  generation, evolution and learning}.
\newblock \bibinfo{journal}{Australian Journal of Telecommunications and the
  Digital Economy} \bibinfo{volume}{6}, \bibinfo{pages}{35--58}.
\bibitem[{Smith et~al.(2018b)Smith, Hunjet and Khan}]{smith2018swarm}
\bibinfo{author}{Smith, P.}, \bibinfo{author}{Hunjet, R.},
  \bibinfo{author}{Khan, A.}, \bibinfo{year}{2018}b.
\newblock \bibinfo{title}{Swarm learning in restricted environments: an
  examination of semi-stochastic action selection}, in:
  \bibinfo{booktitle}{2018 15th International Conference on Control,
  Automation, Robotics and Vision (ICARCV)}, \bibinfo{organization}{IEEE}. pp.
  \bibinfo{pages}{848--855}.
\bibitem[{Smith-Miles(2008)}]{smith2008towards}
\bibinfo{author}{Smith-Miles, K.A.}, \bibinfo{year}{2008}.
\newblock \bibinfo{title}{Towards insightful algorithm selection for
  optimisation using meta-learning concepts}, in: \bibinfo{booktitle}{Neural
  Networks, 2008. IJCNN 2008.(IEEE World Congress on Computational
  Intelligence). IEEE International Joint Conference on},
  \bibinfo{organization}{IEEE}. pp. \bibinfo{pages}{4118--4124}.
\bibitem[{Soria-Alcaraz et~al.(2014)Soria-Alcaraz, Ochoa, Swan, Carpio, Puga
  and Burke}]{soria2014}
\bibinfo{author}{Soria-Alcaraz, J.A.}, \bibinfo{author}{Ochoa, G.},
  \bibinfo{author}{Swan, J.}, \bibinfo{author}{Carpio, M.},
  \bibinfo{author}{Puga, H.}, \bibinfo{author}{Burke, E.K.},
  \bibinfo{year}{2014}.
\newblock \bibinfo{title}{Effective learning hyper-heuristics for the course
  timetabling problem}.
\newblock \bibinfo{journal}{European Journal of Operational Research}
  \bibinfo{volume}{238}, \bibinfo{pages}{77--86}.
\bibitem[{Tabataba and Mousavi(2012)}]{tabataba2012}
\bibinfo{author}{Tabataba, F.S.}, \bibinfo{author}{Mousavi, S.R.},
  \bibinfo{year}{2012}.
\newblock \bibinfo{title}{A hyper-heuristic for the longest common subsequence
  problem}.
\newblock \bibinfo{journal}{Computational biology and chemistry}
  \bibinfo{volume}{36}, \bibinfo{pages}{42--54}.
\bibitem[{Tavares et~al.(2018)Tavares, Anbalagan, Marcolino and
  Chaimowicz}]{tavares2018algorithms}
\bibinfo{author}{Tavares, A.R.}, \bibinfo{author}{Anbalagan, S.},
  \bibinfo{author}{Marcolino, L.S.}, \bibinfo{author}{Chaimowicz, L.},
  \bibinfo{year}{2018}.
\newblock \bibinfo{title}{Algorithms or actions? a study in large-scale
  reinforcement learning.}, in: \bibinfo{booktitle}{IJCAI}, pp.
  \bibinfo{pages}{2717--2723}.
\bibitem[{Terashima-Mar{\'{i}}n et~al.(2008)Terashima-Mar{\'{i}}n,
  Ortiz-Bayliss, Ross and Valenzuela-Rend{\'o}n}]{terashima2008}
\bibinfo{author}{Terashima-Mar{\'{i}}n, H.}, \bibinfo{author}{Ortiz-Bayliss,
  J.C.}, \bibinfo{author}{Ross, P.}, \bibinfo{author}{Valenzuela-Rend{\'o}n,
  M.}, \bibinfo{year}{2008}.
\newblock \bibinfo{title}{Hyper-heuristics for the dynamic variable ordering in
  constraint satisfaction problems}, in: \bibinfo{booktitle}{Proceedings of the
  10th annual conference on Genetic and evolutionary computation},
  \bibinfo{organization}{ACM}. pp. \bibinfo{pages}{571--578}.
\bibitem[{Thabtah and Cowling(2008)}]{Thabtah2008}
\bibinfo{author}{Thabtah, F.}, \bibinfo{author}{Cowling, P.},
  \bibinfo{year}{2008}.
\newblock \bibinfo{title}{Mining the data from a hyperheuristic approach using
  associative classification}.
\newblock \bibinfo{journal}{Expert Systems with Applications}
  \bibinfo{volume}{34}, \bibinfo{pages}{1093--1101}.
\newblock \DOIprefix\doi{10.1016/j.eswa.2006.12.018}.
\bibitem[{Trianni et~al.(2016)Trianni, De~Simone, Reina and
  Baronchelli}]{trianni2016emergence}
\bibinfo{author}{Trianni, V.}, \bibinfo{author}{De~Simone, D.},
  \bibinfo{author}{Reina, A.}, \bibinfo{author}{Baronchelli, A.},
  \bibinfo{year}{2016}.
\newblock \bibinfo{title}{Emergence of consensus in a multi-robot network: From
  abstract models to empirical validation}.
\newblock \bibinfo{journal}{IEEE Robotics and Automation Letters}
  \bibinfo{volume}{1}, \bibinfo{pages}{348--353}.
\bibitem[{Zhao et~al.(2004)Zhao, Ammar and Zegura}]{zhao2004}
\bibinfo{author}{Zhao, W.}, \bibinfo{author}{Ammar, M.},
  \bibinfo{author}{Zegura, E.}, \bibinfo{year}{2004}.
\newblock \bibinfo{title}{A message ferrying approach for data delivery in
  sparse mobile ad hoc networks}, in: \bibinfo{booktitle}{Proceedings of the
  5th ACM international symposium on Mobile ad hoc networking and computing},
  \bibinfo{organization}{ACM}. pp. \bibinfo{pages}{187--198}.

\end{thebibliography}
	\appendix

\section{RHGN Pattern} 

\textbf{Networking}
\begin{itemize} 
	\item neighbourhood size
	\item sink ID
	\item source ID
	\item  unique sinks in 10 steps
	\item sink changes in 10 steps
	\item unique sources in 10 steps
	\item source changes in 10 steps
	\item unique sinks in 100 steps
	\item sink changes in 100 steps
	\item unique sources in 100 steps
	\item source changes in 100 steps
	\item unique sinks in 500 steps
	\item sink changes in 500 steps
	\item unique sources in 500 steps
	\item source changes in 500 steps
	\item unique sinks in 1000 steps
	\item sink changes in 1000 steps
	\item unique sources in 1000 steps
	\item source changes in 1000 steps
 	\item jamming strength
	\item network noise state ($<-95dB$, $-95:-85.5dB$, $-85.5:-71.25dB$, $-71.25:-47.5dB$, $-47.5:-23.75dB$, $-23.75:0dB, >0dB$) 
	)
\end{itemize}  
  
  \textbf{Packets}
  \begin{itemize} 
\item agent packets held
\item closest swarm neighbour has $\ge 1 $ packets
\item closest swarm neighbour is packet full
\item sink-ward closest swarm neighbour has $\ge 1 $ packets
\item sink-ward closest swarm neighbour is packet full
\item source-ward closest swarm neighbour has $\ge 1 $ packets
\item source-ward closest swarm neighbour is packet full
\item closest non-swarm neighbour has $\ge 1 $ packets
\item closest non-swarm neighbour is packet full
\item sink-ward closest non-swarm neighbour has $\ge 1 $ packets
\item sink-ward closest non-swarm neighbour is packet full
\item source-ward closest non-swarm neighbour has $\ge 1 $ packets
\item source-ward closest non-swarm neighbour is packet full
\end{itemize}

\textbf{Distance}
  \begin{itemize} 
\item  assumed source to sink distance (rounded to 1 dec.)
\item distance of closest swarm neighbour (rounded to 1 dec.)
\item signal strength of closest swarmneighbour (rounded to 1 dec.)
\item distance of sink-ward closest swarm neighbour (rounded to 1 dec.)
\item signal strength of sink-ward closest swarm neighbour (rounded to 1 dec.)
\item distance of source-ward closest swarm neighbour (rounded to 1 dec.)
\item signal strength of source-ward closest swarm neighbour (rounded to 1 dec.)
\item distance of roughly sink-ward closest swarm neighbour (rounded to 1 dec.)
\item distance of closest non-swarm neighbour (rounded to 1 dec.)
\item signal strength of closest non-swarmneighbour (rounded to 1 dec.)
\item distance of sink-ward closest non-swarm neighbour (rounded to 1 dec.)
\item signal strength of sink-ward closest non-swarm neighbour (rounded to 1 dec.)
\item distance of source-ward closest non-swarm neighbour (rounded to 1 dec.)
\item signal strength of source-ward closest non-swarm neighbour (rounded to 1 dec.)
\item closest wall distance (rounded to 1 dec.)

  \end{itemize}
  \label{pattern}

\end{document}